%% file: acl_latex.tex
\pdfoutput=1

\documentclass[11pt]{article}

\usepackage[]{acl}
\usepackage{booktabs}    
\usepackage{multicol}
\usepackage{multirow}
\usepackage{colortbl} 
 
\usepackage{times}
\usepackage{latexsym} 
\usepackage{amsmath, amssymb}

\usepackage{textcomp}

\usepackage{pifont}
\usepackage{longtable}
\usepackage{makecell}
\usepackage[T1]{fontenc}

\usepackage[utf8]{inputenc}

\usepackage{microtype}

\usepackage{array}
\usepackage{inconsolata}

\usepackage{graphicx}

\title{Reverse Preference Optimization for Complex Instruction Following}

\author{
Xiang Huang\textsuperscript{1,2}, Ting-En Lin\textsuperscript{1},  Feiteng Fang\textsuperscript{1}, Yuchuan Wu\textsuperscript{1}, \\ 
\textbf{Hangyu Li\textsuperscript{1}, Yuzhong Qu\textsuperscript{2}\footnotemark[1] , Fei Huang\textsuperscript{1}, Yongbin Li\textsuperscript{1}\thanks{Corresponding authors.}} \\ 
\textsuperscript{1}Tongyi Lab \\
\textsuperscript{2}State Key Laboratory for Novel Software Technology, Nanjing University, China \\
  \texttt{\{chengjun.hx, ting-en.lte, shengxiu.wyc, shuide.lyb\}@alibaba-inc.com}}
 
\begin{document}
\maketitle
\begin{abstract}
Instruction following (IF) is a critical capability for large language models (LLMs). However, handling complex instructions with multiple constraints remains challenging. 
Previous methods typically select preference pairs based on the number of constraints they satisfy, introducing noise where chosen examples may fail to follow some constraints and rejected examples may excel in certain respects over the chosen ones.  
To address the challenge of aligning with multiple preferences, we propose a simple yet effective method called Reverse Preference Optimization (RPO). 
It mitigates noise in preference pairs by dynamically reversing the constraints within the instruction to ensure the chosen response is perfect, alleviating the burden of extensive sampling and filtering to collect perfect responses. 
Besides, reversal also enlarges the gap between chosen and rejected responses, thereby clarifying the optimization direction and making it more robust to noise. 
We evaluate RPO on two multi-turn IF benchmarks, Sysbench and Multi-IF, 
demonstrating average improvements over the DPO baseline of 4.6 and 2.5 points~(on \textit{Llama-3.1 8B}), respectively. 
Moreover, RPO scales effectively across model sizes (8B to 70B parameters), with the 70B RPO model surpassing \textit{GPT-4o}.  

\end{abstract}

\section{Introduction} 
Large Language Models (LLMs)~\cite{brown2023language,openai2023gpt4, tao2024survey} have demonstrated impressive capabilities in various tasks such as code, math, and text generation.  
Among all abilities, instruction following~\cite{zhou2023instruction,he2024multi} has emerged as a key ability for conversational intelligence as it reflects how well LLMs can align with human preferences.
In a real-world scenario, user intention typically involves diverse and multiple preferences, posing a challenge to accurately meet the expectations of various users.

\begin{figure}[t]
    \centering    
    \includegraphics[width=\linewidth]{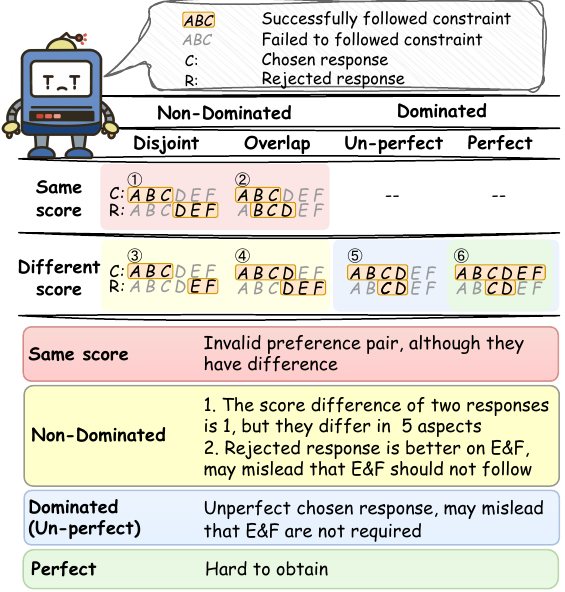}
    \caption{Noise in a multi-preference scenario. 
    Any two responses that differ in at least one aspect belong to the above six types (case \textcircled{\footnotesize{1}} to \textcircled{\footnotesize{6}}).
    ABCDEF are six constraints that need to be followed, and the color in gray indicates this response failed to follow them.     
    C: and R: represent the adherence of chosen and rejected responses to different constraints, respectively.
    }
    \label{fig:issue_example}
\end{figure}

However, as the number of preferences that need to be followed increases, aligning multiple preferences simultaneously becomes a challenging problem. 
Previous methods typically construct preference pairs for alignment
and distinguish chosen and rejected responses based on their total scores~(the number of constraints that were successfully followed).  
In a multi-preference alignment scenario, this method presents two main drawbacks:
 
Firstly, using the total score difference as the gap between two responses may not accurately reflect the true differences between them, potentially underestimating their variations.
Taking Case \textcircled{\footnotesize{1}} in Figure \ref{fig:issue_example} as an example, although the two responses have the same total score, they exhibit differences in adherence to 6 constraints.
 
Secondly, selecting preference pairs in a compositional setting may introduce noise. 
In Cases \textcircled{\footnotesize{3}} and \textcircled{\footnotesize{4}}, both chosen and rejected responses have aspects where they outperform the other. 
Although the chosen response has a higher total score, using it directly as a preference pair may mislead the model to assume that constraints E and F should not be followed, as the chosen response does not satisfy them, while the rejected response does.

We notice that the above issues can be mitigated if the chosen response is perfect~(Case \textcircled{\footnotesize{6}}).
However, perfect responses are hard to obtain. 
It requires a large amount of sampling or refinement, and it may not be feasible to sample a perfect response for a weak model.
Besides, pursuing a perfect response is not an economical and scalable solution.
As the number or difficulty of constraints increases, the difficulty of sampling a perfect response increases dramatically, and there are always cases where perfect responses can not be obtained in a limited sample time.
 
To address these issues, we propose \textbf{R}everse \textbf{P}reference \textbf{O}ptimization~(RPO) to efficiently align with multiple preferences. 
The idea of RPO emerged from two key observations:
1). Learning a constraint has no fundamental difference with learning its opposite.
2). For a clear and unambiguous constraint, a response either adheres to it or deviates from it.
Specifically, RPO dynamically reverses the constraints within the instructions to ensure that the chosen response fully meets all constraints.
In this way, any response can be easily transformed into perfect ones for the new instruction, eliminating labor-intensive sampling and filtering to pursue perfect responses.
Besides, it also amplifies the gap between chosen and rejected responses, thus clarifying the optimization direction and enhancing robustness to noise. 

We summarize our contributions as follows: 
\begin{itemize}
    \item We systematically analyzed the noise issues in multi-preference alignment and introduced RPO, which eliminates noise by reversing constraints, thereby reducing the burden of data collection and revealing the real difference of response pairs.
    \item We developed a role-driven self-play framework to produce natural and diverse multi-turn IF data and constructed a corpus called SysBank, which consists of 30K system prompts originating from real scenarios.
    \item We conducted extensive experiments on models of various sizes and series. On Llama 8B base model, RPO significantly outperforms DPO baseline by 4.6 and 2.5 on two multi-turn complex IF datasets. 
    Besides, RPO also surpasses \textit{GPT-4o} with a 70B model.
\end{itemize}

\section{Related Work}

\subsection{Instruction Following Evaluation}
Instruction following evaluates the ability to follow instructions with certain constraints, such as ``write a story about Ne Zha with \textbf{less than 200 words} and \textbf{end it with an emoji}''.
\citet{zhou2023instruction} first propose this challenge and construct IFEVAL dataset with verifiable constraints that can be judged with simple rules.
Subsequent benchmarks such as FollowBench~\cite{jiang2024followbench} and CFBench~\cite{zhang2024cfbench} focus on more complex instruction which requires to satisfy multiple constraints.
ComplexBench~\cite{wen2024benchmarking} explores the composition structure among different constraints, such as And, Chain, and Selection.
SysBench~\cite{qin2024sysbench} further analyzes how well the model can follow system messages with multiple preferences in a multi-turn setting, which is closer to downstream Agent applications, such as assistant chatbot or role-playing.
MultiIF~\cite{he2024multi} also focuses on the multi-turn settings by extending IFEVAL to three-turn conversations.
In this paper, we mainly focus on improving the ability to follow multi-constraint in a multi-turn conversation setting.

\subsection{Instruction Following Methods}
Given that instruction following fundamentally involves aligning with human preferences, most of the works have adopted DPO-based~(Direct Preference Optimization) methods to tackle this task~\cite{he2024complex,qin2024infobench,sun2024conifer, liu2025survey}.
Some research further employs the Online-DPO framework:
AutoIF~\cite{dong2024self} leverages LLMs to generate validation code for automatically assessing data quality, and AutoDetect~\cite{cheng2024autodetect} iteratively uncovers and rectifies model shortcomings. 
SPAR~\cite{cheng2024spar} integrates a tree-search self-refinement process to suppress noise, steering the model’s attention toward satisfying constraints. 
Another line of work emphasizes the value of the instructions themselves: 
IOPO~\cite{zhang2024iopo} treats the perfect responses of two different instructions as the rejected sample for each other, thereby aligning both input and output preference pairs. 
CRAB~\cite{qi2024constraint} uses back-translation to generate instruction based on the response to improve the quality of training data.
In this paper, we propose a simple method to efficiently produce noise-free preference data by reversing the constraint within the instruction.
It mitigates the issue of noise interference and alleviates the burden to sample perfect response.

\section{Preliminary} 

\noindent\textbf{Reinforcement Learning from Human Feedback~(RLHF)}~\cite{ouyang2022rlhf}
The goal of traditional RLHF encompasses two primary goals:  maximizing the rewards of responses and minimizing the deviation~(KL divergence) of the aligned policy model $\pi_{\theta}$ from the initial reference (SFT) model $\pi_{ref}$.
Therefore, this RL objective can be formulated as:
\begin{equation}\label{eq:RL}
\small
\begin{split}
\pi_{\theta}(y|x) = \max_{\pi_{\theta}}\ & \mathbb{E}_{{x}\sim\mathcal{D}, {y}\sim \pi_{\theta}(y \mid x)}  \bigl[r(x, y)] \\- 
\beta\mathbb{D}_{\textrm{KL}}&\bigl[\pi_{\theta}(y\mid x)\mid \mid \pi_{ref}(y\mid x)\bigr]
\end{split}
\end{equation}
where $x$ is the instruction, and $y$ indicate the response sampled from $\pi_{\theta}$.
The reward model $r$ is trained from preference pairs dataset $\mathcal{D}$
and is used to provide feedback to the language model.
Let $y_w$ and $y_l$ be the preferred and undesired responses, the optimization objective of $r$ is formulated as:
\begin{equation}\label{eq:reward_model}
\small
    \mathcal{L}_r = -\mathbb{E}_{(x, y_w, y_l)\sim \mathcal{D}}\bigl[\log \sigma(r(x, y_w)- r(x, y_l))\bigr]
\end{equation}
\noindent\textbf{Direct Preference Optimization~(DPO)} \cite{rafailov2024direct} re-parameter the RL object and build the connection between the reward model $r$ and the target policy model $\pi_{\theta}$.
By incorporating the reward model into the Bradley-Terry~(BT) ranking objective~\cite{Bradley1952RankAO} $p(y_w \succ y_l~|~x) = \sigma(r(x,y_w)-r(x,y_l))$, DPO aims to maximize the probability of the preferred output $y_{w}$ and minimize that of the undesirable output $y_{l}$. 
The optimization objective is formulated as:
\begin{equation} 
\small
\begin{aligned}
    &\mathcal{L}_\text{DPO} = -\mathbb{E}_{(x, y_w, y_l)\sim \mathcal{D}} \\&\Bigg[\log \sigma \Bigg(\beta \log \frac{\pi_{\theta}(y_w\mid x)}{\pi_{ref}(y_w\mid x)} \quad-\beta \log \frac{\pi_{\theta}(y_l\mid x)}{\pi_{ref}(y_l\mid x)}\Bigg)\Bigg]
\end{aligned}
\end{equation}

\label{sec:method}
\begin{figure*}[ht!]
    \centering
    \includegraphics[width=\linewidth]{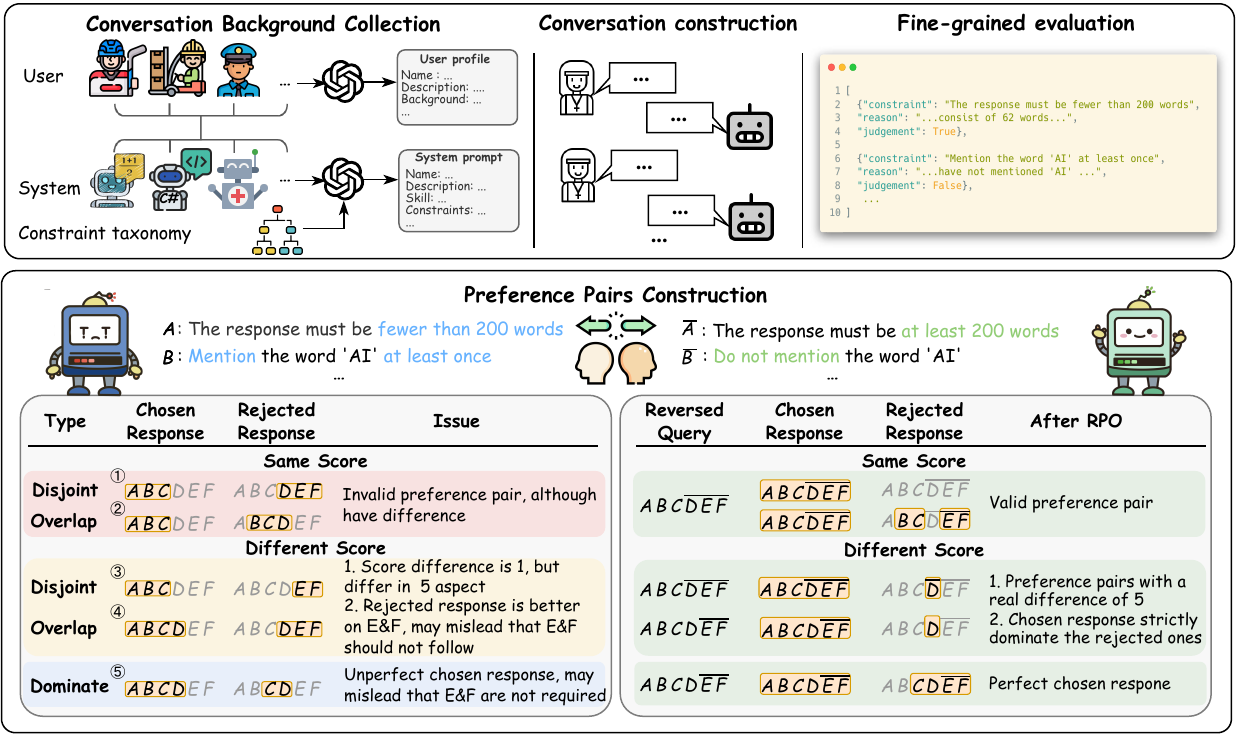}
    \caption{Illustration of RPO and data construction.
    We first model various users and systems and then engage them in self-play. For responses sampled during the dialogue, we perform fine-grained evaluations to assess adherence to each constraint. 
    Different types of response pairs are converted into noise-free preference pairs, as depicted in the lower part of the figure.}
    \label{fig:RPO}
\end{figure*}

\section{Reverse Preference Optimization}
\label{subsec:RPO} 
\subsection{Objective}
Let $x$ be the instruction with constraints, and $y$ indicate the response sampled from models $\pi_{\theta}$. 
$\mathcal{S}_{y}$ is a set of boolean value indicating whether $y$ follow each constraint in $x$.
$y_i$ and $y_j$ are two responses that differ in adherence to at least one constraint, denoted as $\mathcal{S}_{y_i} \neq \mathcal{S}_{y_j}$. 
$x_{\mathcal{S}_i}$ represent the instruction after reverse the constraint which $y_i$ failed to satisfied.
For $x_{\mathcal{S}_i}$, $y_i$ is a perfect response and is strictly dominates $y_j$, denoted as $y_i \succ y_j$.
We also introduced an adaptive margin $\gamma g$ to alleviate the issue of uniformly handling preferences with different gaps. 
$g$ is the number of differences between $\mathcal{S}_{y_i}$ and $\mathcal{S}_{y_j}$, ranging from $0$ to $|\mathcal{S}_x|$.
In this way, the optimization objective of RPO is formulated as: 
\begin{equation} 
\small
\begin{aligned} 
    &\mathcal{L}_\text{RPO} = -\mathbb{E}_{(x_{S_i},  y_i, y_j, g)\sim \mathcal{D}}\\&\left[\log \sigma  \left(  \beta \log \frac{\pi_{\theta}(y_i|x_{S_i})}{\pi_{ref}(y_i|x_{S_i})} - \right.\right. \left.\left.\vphantom{\frac{\pi_{\theta}}{\pi_{ref}}}\beta \log \frac{\pi_{\theta}(y_j|x_{S_i})}{\pi_{ref}(y_j|x_{S_i})} -  \gamma g \right) \right]
\end{aligned}
\end{equation}
where $\gamma,\beta$ are hyper-parameters.

\subsection{Difference Measurement}
Instead of using the total score difference to measure the degree of the difference between responses, 
we the number of constraints where the two responses differ in their adherence.  
If the two responses separately follow a constraint and do not follow it, their difference on this constraint is 1; otherwise is 0.
Two responses can be used as a valid preference pair if they have differences in following any constraint, even if they achieve the same total scores.
For example, in Figure 1, two responses of Case 1 follow constraints [A, B, C] and [D, E, F], respectively.
They have identical total scores but represent completely different behaviors with a true difference of 6 points.

\subsection{Reverse constraint}
We reverse constraints that a response fails to satisfy, transforming it into a perfect response to serve as the chosen response.  
For the Case \textcircled{\footnotesize{4}} in Figure~\ref{fig:RPO},
where the two responses respectively follow [A,B,C,D] and [D,E,F],  treating the first response as a chosen response may mislead the model that following constraints E and F are undesirable as the chosen response does not follow them while the rejected response does. 
We revert constraints E and F in the system prompt to their opposite.
Consequently, the chosen response dominates the rejected ones and performs no worse than the rejected response in any aspect, thus eliminating noise.
A reverse example is shown in the middle of Figure~\ref{fig:RPO}: assuming constraint $A$ represents ``The response must be fewer than 200 words'', its opposite constraint $\overline{A}$ is ``... at least 200 word''.
This reversal process is a relatively simple task, and prompting LLM can yield high-quality results.
Apart from reducing noise, reversal also amplifies the gap between chosen and rejected responses, thereby clarifying the optimization direction and enhancing robustness against noise introduced by evaluation errors. 
In the above case, the score difference rises from 1 to 5, accurately reflecting the real difference between the responses.
To summarise, RPO has the following advantages:
\begin{itemize}
    \item Noise-free: The chosen responses are perfect and strictly superior to rejected ones.
    \item Sample-effectiveness: RPO does not seek to sample perfect responses as the chosen ones; it only needs two responses that have differences in adhering to any constraints.  
    \item  Simple and precise: Reversal is simpler than sampling perfect responses, self-refine, or back-translation, which also contributes to a high success rate.
\end{itemize}


\section{Data Collection}

\subsection{Overview} 
As previous methods primarily focus on enhancing the single-turn IF ability, they typically sample constraints from a constraint pool and attach them to existing instruction datasets. However, we argue that this method suffers from several limitations. 
Firstly, it lacks diversity.
The constraints and instruction data are general and not tailored for specific downstream scenarios.
As a result, they exhibit considerable homogeneity, and LLMs may have already encountered these instructions during pre-training. 
Secondly, this method is inadequate for constructing multi-turn dialogue data. 
Merely concatenating single-turn instructions is hard to produce a coherent conversation.
Moreover, attaching constraints to pre-existing instructions might lead to unnatural adaptation and introduce an extra burden for validation to ensure compatibility.

To alleviate the above challenges, inspired by \citet{ge2024scaling}, we employ a role-driven self-play framework to generate more diverse data in a multi-turn setting.  
We performed role modeling for users and the system, assigning them with unique profiles.
By leveraging these profiles as a context-specific conversation background, we introduce variations during generation, thereby preventing the repeated generation of similar instructions.
The introduction of role can be seen as an inspired corpus~\cite{xu2023wizardlm} to enrichh diversity.
Theoretically, combinations of different systems and users can produce entirely distinct conversations, resulting in a vast amount of diverse instruction-following data.
The following data constructions are all sampled from \textit{GPT-4o-mini}, including profile, conversation generation, and quality evaluation. 
The responses that are used to construct preference pairs are sampled from base model~(\textit{Llama-3.1 8B Instruct}). 

\subsection{Conversation Background Collection}

\noindent\textbf{Constraint Collection.} 
Instead of self-evol on some constraint examples and samples from the expanded constraints pool, we construct a comprehensive constraint taxonomy. 
It can serve as a high-level awareness for LLM to understand what a constraint is. 
Note that the constraint is part of the system prompt.
Actually, we provide LLMs with the simple role description of this system and the constraint ontology to generate system profiles and system constraints concurrently.
It simplifies the process and ensures that the constraints can naturally adapted to system identity. 
More detail can be found in Appendix~\ref{appdix:constraint_ontology_prompt}.

\noindent\textbf{Profile Collection.}
For system-side role modeling, we curated real user-written GPTs data, from BlackFriday-GPTs\footnote{https://github.com/friuns2/BlackFriday-GPTs-Prompts/tree/main/} and GPTstore\footnote{https://gptstore.ai/}. 
We filtered out data unrelated to pure text, such as audio, image, or video-related assistants.
As GPTstore only provides access to a system name and simple description,
we then expanded them to comprehensive profiles with constraints, resulting in 30K system prompts, which we called \textbf{SysBank}.
To the best of our knowledge, SysBank is the first large-scale system prompt dataset that originates from diverse real system roles and is equipped with rich constraints.
On the user side, we expanded the profile from personas~\cite{ge2024scaling}, which includes simple descriptions of 200K distinct roles, to create vivid and detailed user profiles.  
We use a 2.5K subset of these two corpus to construct subsequent training data and will release the complete corpus.
The prompts for generating system and user profiles are detailed in Appendix \ref{appdix:profile_prompt}.

\subsection{Conversation Construction}
\label{subsec:conversation_construction} 
We facilitate the system and user to engage in self-play, generating dialogue histories up to five turns. 
For query generation, we provide LLMs with 
the constraints that the system needs to follow, 
along with the conversation history, 
to generate challenging and continuity questions. 
In constructing the responses, 
we employ two strategies to enhance data quality. 
First, we appended the constraints to follow at the end of the user query. 
This is to prevent the model from ignoring the initial system prompt as the turn of conversation history increases. 
Additionally, we conduct up to three times self-refine for responses with a following rate of less than 0.8 and terminate this session if the response is still not good enough after self-refine.
The conversation history data is used to train the SFT model and construct a unified dialogue history for a fair comparison of different alignment methods. 
\subsection{Quality Evaluation Method} 
To evaluate the quality of the sampled responses, we use LLMs to determine whether the response adheres to each constraint one by one.
Considering the autoregressive nature of LLMs, for each constraint, 
we require LLMs first to repeat the original constraint, followed by an explanation of why it considers this response follow or not follow to the constraint, and finally provide the overall judgment~(example in Figure~\ref{fig:RPO}). 
Repeating the original constraint helps reinforce the LLM's impression of the constraint while concluding with the judgment prevents the LLM from directly delivering an incorrect result without careful consideration.
More detail can be found in Appendix~\ref{appdix:evaluation_prompt}

\subsection{Preference Pair Construction} 
We sample five times for each turn to obtain responses.
For any two responses that exhibit any difference in adherence to the constraints, we transform both of them into the perfect response using the method in Section~\ref{subsec:RPO} and separately treat them as chosen examples to form two preference pairs.  
In this way, for constraints that originally contribute to noise (where one response follows it and the other does not), the constructed two preference pairs allow the model to learn both how to follow the original constraint and its opposite.

\begin{table*}[t]
    \centering
    \resizebox{0.95\textwidth}{!}{
    \begin{tabular}{>{\raggedright\arraybackslash}p{4.5cm} >{\raggedleft\arraybackslash}p{1.2cm} >{\raggedleft\arraybackslash}p{1.2cm} >{\raggedleft\arraybackslash}p{1.2cm} >{\raggedleft\arraybackslash}p{1.2cm} >{\raggedleft\arraybackslash}p{1.2cm} >{\raggedleft\arraybackslash}p{1.2cm} >{\raggedleft\arraybackslash}p{1.2cm} >{\raggedleft\arraybackslash}p{1.2cm}}
    \toprule
    \multirow{2}{*}{\textbf{Method}} &  \multicolumn{4}{c}{\textbf{Sysbench}}  & \multicolumn{4}{c}{\textbf{Multi-IF}} \\ 
    \cmidrule(lr){2-5} \cmidrule(lr){6-9}
    &  \textbf{CSR} & \textbf{ISR} & \textbf{SSR} & \textbf{Avg.}   & \textbf{Step 1} & \textbf{Step 2} & \textbf{Step 3}  & \textbf{Avg.} \\ 
         \midrule        
       GPT-3.5-Turbo &  75.51 & 61.51 & 33.44   & 56.82 & 72.11	& 56.59	& 47.41  &  58.70 \\
       GPT-4o-mini & 86.60 & 77.32 & 54.24  &  72.72 & 83.58	& 73.58	& 65.23   &  74.13 \\
       GPT-4o & 89.72 & 81.71 & 61.51    & 77.65 & \textbf{84.70} & \textbf{76.00} & 68.33  &   \textbf{76.34} \\ 
       Claude-3.5 Sonnet  & \textbf{94.64} & \textbf{89.68} & \textbf{74.36}  & \textbf{86.23} & 83.87 &	74.87 &	\textbf{69.80}   & 76.18 \\ 
       \midrule
       Llama-3.1 8B Instruct  & 71.89 & 55.87 & 28.59   & 52.12 & 74.69 & 67.00 & 57.99  &  66.56 \\
       Llama-3.1 8B Instruct SFT &75.74 & 60.82 & 32.69  & 56.42 & 71.31 & 62.56 & 55.62  &  63.16 \\
       Llama-3.1 8B Instruct DPO & 80.56 & 66.67 & 40.37 & 62.53 &   74.57 & 66.90 &58.50  &  66.66 \\
       Llama-3.1 8B Instruct   KTO  & 80.41 & 68.15 & 41.22 &  63.26 &  77.39 & 69.47 & 59.62   &  68.83 \\ 
       Llama-3.1-8B Instruct  RPO & \textbf{83.10}	& \textbf{71.27}	& \textbf{46.99}   & \textbf{67.12}   &\textbf{77.50} & \textbf{69.47} & \textbf{60.57}   & \textbf{69.18} \\  
       \midrule 
       Llama-3.1 70B Instruct & 81.34 &	68.11 & 42.17  & 63.87& 83.69 & 75.13 & 67.89   & 75.57 \\
       Llama-3.1 70B Instruct SFT  & 79.12 & 65.70 &	38.07  & 60.96
 & 82.54 & 72.83 & 65.04  &  73.47 \\
       Llama-3.1 70B Instruct DPO  & 85.91 & 75.12 & 50.89   &  70.64
 & 84.37 & 75.63 & 67.76   & 75.92 \\ 
       Llama-3.1 70B Instruct KTO   & 84.69 & 73.92 & 48.37    & 68.99 & 83.53 & 75.76 & 67.62   & 75.64 \\
       Llama-3.1 70B Instruct RPO  & \textbf{89.54} & \textbf{81.76} & \textbf{62.20}  &  \textbf{77.83} & \textbf{86.47} & \textbf{77.77} & \textbf{70.21}   & \textbf{78.15} \\ 
      \midrule
       Qwen-2.5 7B Instruct  & 73.14 & 56.57 & 31.06     & 53.59 & 76.43 &  61.97 &  51.20  &  63.20 \\
       Qwen-2.5 7B Instruct SFT & 70.64 & 54.66 & 27.90     & 51.07 & 73.38 & 58.08 &  48.53    &   60.00 \\ 
       Qwen-2.5 7B Instruct DPO  & 77.64 & 63.06 & 37.76     & 59.49 & 76.10  & 62.08  &  51.16     & 63.11 \\       
       Qwen-2.5 7B Instruct   KTO  & 78.22 & 64.21 & 37.46   &  59.96
 & 75.08 & 62.55 & 52.88      & 63.50 \\
       Qwen-2.5 7B Instruct  RPO &  \textbf{80.25}	& \textbf{67.16}	& \textbf{41.77}     & \textbf{63.06} &  \textbf{77.25}  & \textbf{65.43} & \textbf{56.12}   & \textbf{66.27} \\
       \midrule
       Qwen-2.5 72B Instruct  & 82.33 & 70.20 & 47.40   & 66.64 &  86.13 & 74.92 & 64.39   & 75.15 \\
       Qwen-2.5 72B Instruct SFT   &  85.56 & 74.18  & 50.45  & 70.06 &  85.73& 73.00 & 64.32  &  74.35 \\
       Qwen-2.5 72B Instruct DPO   & 88.42 & 	79.18	 & 59.92   &  75.84
   & 85.37 & 75.61 & 67.31  &  76.10 \\ 
       Qwen-2.5 72B Instruct KTO   & 87.79 & 78.99 & 55.90   & 74.23 &   81.38 & 71.55  & 58.03   & 70.32 \\
       Qwen-2.5 72B Instruct RPO  & \textbf{89.48} & \textbf{80.98} & \textbf{60.44}   & \textbf{76.97} &   \textbf{86.92} & \textbf{78.04} & \textbf{70.08} & \textbf{78.35} \\
     \bottomrule
    \end{tabular}
    }
    \caption{Main result of two multi-turn complex instructions following dataset. 
    For SysBench, we report CSR, ISR, and SSR.    
    For Multi-IF, we report the average accuracy in three steps.
    The accuracy are the average of four metrics: Prompt-level strict-accuracy, Inst-level strict-accuracy, Prompt-level loose-accuracy, Inst-level loose-accuracy}
    \label{tab:main_result}
\end{table*}

\section{Experiment}

\subsection{Setting}
We use Llama-Factory~\cite{zheng2024llamafactory} as the training framework and use Lora~\cite{hu2022lora} and Deepspeed to save memory.
The learning rates for SFT and DPO/KTO/RPO are set to 5e-5 and 5e-4.
All experiments are conducted with 4-8 A100 GPUs.
The default base model is \textit{Llama-3.1-8B-Instruct}.
For each turn, we sample 5 responses to construction preference data for DPO/KTO/RPO.
The $\gamma$ and $\beta$ are set to 0.05 and 0.1.

\subsection{Dataset}

\noindent \textbf{SysBench}~\cite{qin2024sysbench} focus on evaluating the ability to follow system prompts in a multi-turn setting. It consists of 500 five-turn conversations based on six common types of constraints from system messages in real-world scenarios.

\noindent \textbf{MultiIF}~\cite{he2024multi}
expands upon IFEval by extending from single-turn to multi-turn and translating the English prompt into other seven languages, resulting in a dataset of 4,501 multilingual three-turn conversations.
We evaluate the English and Chinese subsets and report their weight average performance.

\subsection{Evaluation Metrics}

\noindent\textbf{CSR} is the average accuracy of constraints satisfied and represents the finest level of granularity.

\noindent\textbf{ISR} indicate the percentage of instructions that satisfied all constraints within an instruction. 

\noindent\textbf{SSR} is a metrics for multi-turn IF. 
It measures the average number of consecutive dialogue turns a model successfully adheres to all constraints starting from the very first turn. 

\noindent\textbf{Average accuracy score}
For Multi-IF, we report the average score of four accuracy scores: prompt-level strict, instruct-level strict, prompt-level loose, and instruct-level loose.
Prompt-level accuracy is 1 when all constraints within a query are satisfied.
Instruct level is the percentage of constraints that are satisfied.
The strict setting is the default evaluation criterion, while the loose setting is computed with a loose criterion following~\cite{zhou2023instruction}.

\subsection{Baseline}
We compare the proposed RPO with Supervised Fine-tuning~(SFT), Directly Preference Optimization~(DPO)~\cite{rafailov2024direct}, and Kahneman-Tversky Optimization~(KTO)~\cite{ethayarajh2024kto} on both 7B and 70B model. 
We also compare with the advanced close-sourced model such as \textit{GPT-4o} and \textit{Claude-3.5 Sonnet}.

\subsection{Main Result}
\noindent\textbf{RPO yield more significant improvement.}
As shown in Table \ref{tab:main_result}, RPO demonstrates significant improvements for both datasets.
On the in-domain benchmark SysBench, RPO demonstrated an average enhancement of over 10 points compared to the Instruct model and significantly outperforms \textit{GPT-3.5-turbo} with an 8B model.
Compared to the DPO baseline, RPO achieves improvements of 2.5, 4.6, and 6.6 points in CSR, ISR, and SSR on \textit{Llama-3.1-8B}. 
On the out-of-domain Multi-IF dataset, RPO similarly exhibits marked improvements. 
Across the three metrics of SysBench, RPO outperforms the DPO baseline by 2.9, 2.6, and 2.1 points, respectively. 

\noindent\textbf{RPO can scale to larger model and adapt to other series models.}
We also report the performance on 70B model and another series model~(e.g., \textit{Qwen-2.5}).
RPO works efficiently on the 70B model and surpasses \textit{GPT-4o} on most metrics.
Besides, experiments on \textit{Qwen-2.5} model exhibit similar superiority to other baselines.

\begin{figure}
    \centering
    \includegraphics[width=0.94\linewidth]{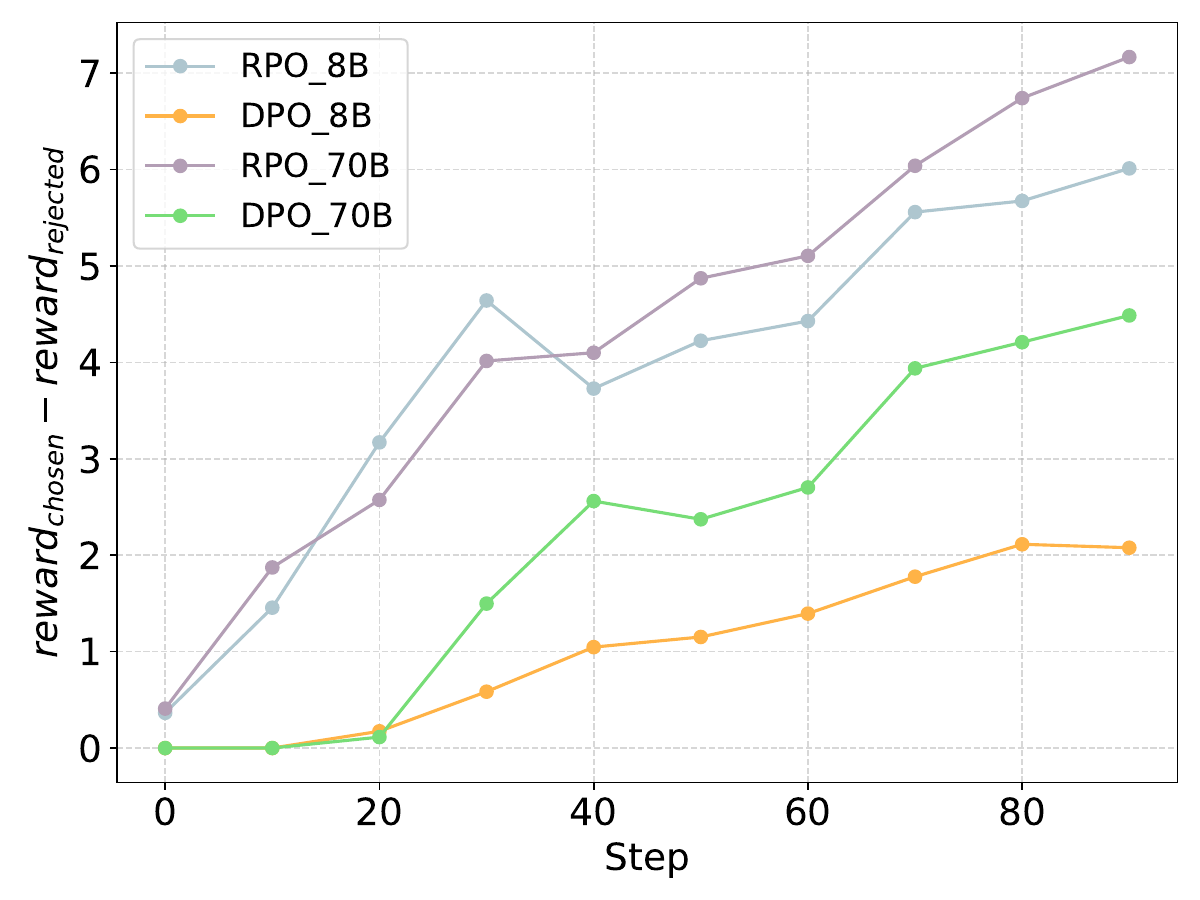}
    \caption{Variation in reward differences between chosen and rejected response over training steps.}
    \label{fig:reward_margin}
\end{figure}

\noindent\textbf{RPO can distinguish chosen and rejected examples more efficiently.}
As the pair-wise alignment method aims to maximize the probability between chosen and rejected examples, optimizing an implicit reward function alongside the policy model, we analyze the trend of the difference between chosen and rejected rewards during the training process of DPO and RPO.
As shown in Figure \ref{fig:reward_margin}, RPO demonstrates a more significant increase in the gap compared to DPO, indicating that RPO is more effective at distinguishing chosen examples from rejected ones.

\subsection{Further Analysis}

\subsubsection{Impact of the chosen-rejected Gap}
As mentioned before, RPO can expose real differences between responses, thereby widening the gap between preference pairs. 
We speculate this can contribute to making the optimization direction clearer, thereby enhancing performance. 
To investigate this, we conducted an experiment using training data with varying score differences, each set containing 5,000 preference pairs. 
As shown in Table \ref{tab:exp_gap}, models trained on easier data consistently outperformed those trained on data with smaller gaps, supporting our hypothesis.
We attribute this to the fact that preference pairs with larger margins are less susceptible to noise and provide clearer optimization guidance. 
Given that the evaluation is based on LLMs, which are inevitably prone to errors, 
training data with larger margins are less likely to be affected by occasional evaluation mistakes and are unlikely to disrupt the order of the preference pairs, thus contributing to more stable and superior performance.

\begin{table}[]
    \centering
    \resizebox{0.49\textwidth}{!}{
    \begin{tabular}{lrrrrrr}
    \toprule
    \multirow{2}{*}{\textbf{Gap}} &  \multicolumn{3}{c}{\textbf{Sysbench}}  & \multicolumn{3}{c}{\textbf{Multi-IF}} \\ 
    \cmidrule(lr){2-4} \cmidrule(lr){5-7} &  \textbf{CSR} & \textbf{ISR} & \textbf{SSR}  & \textbf{Step 1} & \textbf{Step 2} & \textbf{Step 3} \\
    \midrule             
     \textbf{Easy} & \textbf{83.2} & \textbf{69.9} & \textbf{44.1} & \textbf{76.5} & \textbf{68.8} & \textbf{61.3}\\           
     \textbf{Medium}  & 82.9 & 70.3	& 44.2	& 75.5 & 67.3 & 59.8\\      
     \textbf{Hard}  & 80.9 & 67.7 & 41.7 & 75.1 & 66.6 & 59.4\\   
    \bottomrule
    \end{tabular}
    }
    \caption{Performance when training using preference pair with different gaps. Easy, Medium, and Hard set are preference data with gaps of no less than 3, equal to 2, and equal to 1. S1-S3 represents the average accuracy on Step 1-Step 3. 
    }
    \label{tab:exp_gap}
\end{table}

\begin{table}[t]
    \centering
    \resizebox{0.45\textwidth}{!}{
    \begin{tabular}{ccrrr}
    \toprule
    \textbf{Method} & \textbf{Data Size} & \textbf{CSR} & \textbf{ISR} & \textbf{SSR} \\
    \midrule              
    \multirow{3}{*}{\textbf{DPO}} & 5K & 80.56	& 66.67	& 40.37\\
                & 15K & 80.91 &	67.60 &	40.54  \\
                & 20K & 80.25 &	66.80 &	40.41\\
    \midrule
    \multirow{3}{*}{\textbf{RPO}} & 5K & 83.47 & 70.90 & 44.33\\     
                & 15K & 82.07 &	69.84 &	43.48 \\  
                & 20K & 83.10 & 71.27	& 46.99 \\
    \bottomrule
    \end{tabular}
    }
    \caption{Experiment with different training data on SysBench. 
    The base model is Llama 3.1 8B Instruct. 
    }
    \label{tab:same_token}
\end{table}

\subsubsection{Performance under Different Size of Training Data}
Given that RPO can construct more data than DPO under the same response sampling times, this is not a strictly fair comparison with DPO. 
Therefore, we further explored the model performance obtained under different numbers of training data. 
The original training datasets for RPO and DPO are 5K and 20K, respectively. 
We performed down-sampling on RPO data and up-sampling on DPO data. 
The up-sampling was implemented by increasing the response sampling times until we obtained a sufficient number of responses to construct 20K valid preference pairs. 
We also conducted experiments with an intermediate data volume (15k) to strengthen the credibility of the results. 
As shown in Table~\ref{tab:main_result}, across different training data volumes, models trained on RPO data consistently surpassed those trained on DPO data. 
Moreover, neither DPO nor RPO exhibited significant performance differences across varying data volumes, indicating that (1). The original training data volume for DPO in Table \ref{tab:main_result} is sufficient, and (2). RPO can also achieve satisfactory performance with less training data.

\begin{table}[t]
    \centering
    \resizebox{0.45\textwidth}{!}{
    \begin{tabular}{>{\raggedright\arraybackslash}p{1.5cm} >{\raggedleft\arraybackslash}p{1.5cm} >{\raggedleft\arraybackslash}p{1.6cm} >{\raggedleft\arraybackslash}p{1cm} >{\raggedleft\arraybackslash}p{1cm}}
    \toprule
    \multirow{2}{*}{\textbf{Method}}  &  \multirow{2}{*}{\textbf{Valid}} & \multirow{2}{*}{\textbf{Dominated}}  & \multicolumn{2}{c}{\textbf{Perfect}}\\    
    \cmidrule(lr){4-5} & & & <5 & >=5  \\
    \midrule                        
    \textbf{Direct} & 0.77 & 0.63 & 0.59 & 0.23 \\
    \textbf{Refine} & 1.00 & 0.88 & 0.81 & 0.28 \\
    \textbf{Reverse} & 1.00 & 1.00 & 1.00 & 1.00 \\ 
    \bottomrule
    \end{tabular}
    }
    \caption{Sample efficiency analyze.
    We analyze the percentage of different methods that can construct Valid~(have score gap), Dominated~(chosen not worse than rejected on any aspect), and Perfect preference pair~(chosen is perfect).
    We further analyze the Perfect rate when there are fewer than five constraints to follow and no less than five constraints.
    Direct setting constructs preference pairs on these five responses.
    Refine setting additionally performs self-refine for the best response.
    Reverse is the proposed method.
    }
    \label{tab:sample_efficiency}
\end{table}

\subsubsection{Sample Efficiency}

One advantage of RPO is its superior sample efficiency. 
RPO does not strive to sample perfect responses.
In contrast, directly using the originally sampled responses may require numerous sampling attempts to obtain a perfect response.
We analyze the cost associated with obtaining Valid, Dominated, and Perfect preference pairs through direct sampling, refine the best response, and reverse.    
We sample 5 responses for 5,000 queries on \textit{Llama-3.1-8B-Instruct} and refine the best response with \textit{GPT-4o-mini}.
As shown in Table~\ref{tab:sample_efficiency}, when using the original sampled responses to construct preference pairs, only 63\% of the pairs are dominated. 
After incorporating refinement, the quality of the best responses improves, reducing the likelihood that a positive example cannot fully dominate a negative one, but also results in higher costs. 
The proportion of perfect preference pairs is even lower than that of dominant pairs, and this issue becomes increasingly severe as the number of constraints grows.
The perfection rate of responses to instructions with more than five constraints is less than half of that for those with five or fewer constraints.
Conversely, the introduction of the reverse mechanism effectively ensures that any response can be transformed into a perfect response, exhibiting superiority in sample efficiency.

\subsubsection{General Ability}
We also evaluated whether the general capabilities of models aligned on specific tasks degraded, including benchmark about mathematics (GSM8K~\cite{cobbe2021gsm8k}), coding (HumanEval~\cite{chen2021codex}), and alignment (AlignBench~\cite{liu2024alignbench}). 
As shown in Table \ref{tab:general_ability}, except for DPO on GSM8K and SFT on AlignBench, most methods have not exhibited significant degradation. 
Another observation is that larger models tend to incur a smaller alignment tax. For the 70B models, both DPO and RPO introduce bonuses over the Instruct model on some tasks, such as RPO on AlignBench and GSM8K.

\begin{table}[t]
    \centering
    \resizebox{0.45\textwidth}{!}{
    \begin{tabular}{lrrr}
    \toprule
        \textbf{Model} & \textbf{AlignBench}  & \textbf{GSM8K} & \textbf{HumanEval}\\ 
        \midrule        
        \rowcolor{gray!20}\multicolumn{4}{c}{\textit{\textbf{Llama-3.1-8B}}} \\
        Instruct  & 5.34  &  87.49 & 68.29\\
        SFT      & 4.98 & 86.13	& 68.90\\ 
        DPO     & 5.18  & 84.31 & 67.07\\    
        KTO     & 5.37  & 86.73 & 66.46\\    
        RPO          & 5.31  & 88.10 & 68.20 \\
        \midrule    
        \rowcolor{gray!20}\multicolumn{4}{c}{\textit{\textbf{Llama-3.1-70B}}} \\
        Instruct  & 6.15 & 94.92 & 81.71\\
        SFT     & 5.89 & 95.68 & 81.71 \\ 
        DPO     & 5.92  & 95.45 & 79.88\\    
        KTO     & 6.37  & 95.15 & 80.49\\ 
        RPO     & 6.23  & 95.60 & 79.88\\    
        
     \bottomrule
    \end{tabular}
    }
    \caption{General ability of different methods.
    }
    \label{tab:general_ability}
\end{table}
\section{Conclusion}
In this paper, we explore the noise issue of aligning with multiple preferences.
We introduce RPO, a simple and efficient method that dynamically reverses the unsatisfied constraints within the instruction to mitigate noise.
In this way, RPO ensures that the chosen responses are perfect and strictly dominate the rejected ones. 
As RPO does not require a perfect response as the chosen example, it also alleviates the burden of sampling extensive response candidates to construct preference pairs.
Experimental results on two multi-turn instruction-following benchmarks demonstrate that RPO significantly outperforms DPO baseline by 4.6 and 2.5 points on Sysbench and Multi-IF, respectively. 
Besides, RPO can also scale to the 70B model and surpass \textit{GPT-4o} with the 70B model. 
Detailed analysis also shows RPO's superiority in learning efficiency and sample efficiency.
We hope that RPO can serve as a valuable foundation for tasks that require aligning with multiple preferences.

\section*{Limitations} 

Despite the promising results achieved through Reverse Preference Optimization (RPO), there are several limitations to our approach that warrant further exploration.
Firstly, the effectiveness of RPO depends on the quality of the constraints.
Although reversal is a relatively simple task, there are still a small number of bad cases, leading to error propagation in subsequent processes.
Secondly, there exist some constraint that can not be reversed, we have provide a detail discussion in Appendix~\ref{appdix:implicit_constrain} and \ref{appdix:error_analyze}.
Thirdly, there must be at least one preference difference between different responses; otherwise, it is impossible to form a valid RPO preference pair.
In future work, we plan to incorporate RPO into other paradigms, such as online DPO~\cite{guo2024direct} and curriculum learning~\cite{hacohen2019power}.
It is worth evaluating RPO on other LLMs, such as Mixtral~\cite{jiang2024mixtralexperts} and GLM~\cite{du2022glmgenerallanguagemodel}.

\section*{Acknowledgements}
This work was supported by Alibaba Research Intern Program and National Natural Science Foundation of China (NSFC) under Grant No. 62072224. 

\bibliography{custom}
\newpage
\clearpage

\appendix
\input{appendix}

\end{document}

%% file: appendix.tex
\onecolumn

\section{More Discussion of Related Work}
\subsection{Difference with IOPO}
RPO distinguishes itself from IOPO~\cite{zhang2024iopo} by adopting a more streamlined and flexible approach to data handling and constraint management. IOPO relies on constructing two sets of perfect supervised fine-tuning (SFT) data, which is inherently challenging due to the difficulty of achieving perfection. 
Conversely, RPO utilizes random sampling, allowing it to leverage differences between samples without the need for meticulously crafted constraints. 
constraints that are satisfied remain unchanged, while those that are not are simply reversed. 
Moreover, IOPO's approach is limited by a fixed preference score difference of 1, whereas RPO can achieve any desired score difference through the number of constraint reversals. 
This flexibility makes RPO less sensitive to noise, as it does not depend on perfectly evaluated examples to establish a meaningful distinction between positive and negative samples. 
IOPO further complicates the process by requiring the model to propose, modify, or delete constraints, which adds complexity and potential instability. In contrast, RPO's reversal method is straightforward and stable, presenting no intermediate states. 
Additionally, IOPO's process involves multiple intricate steps, including perfect data construction, constraint rewriting, and generating preference pairs. RPO simplifies this by focusing on evaluation, yielding a broader array of preference pairs across various difficulty levels without the need for perfect data construction.

\subsection{Difference with CRAB}
RPO also presents a clear contrast to CRAB~\cite{qi2024constraint}, particularly in its motivation and methodology. CRAB is predicated on the concept of back-translation, leveraging the implicit complex constraints inherently present in existing datasets. 
By generating constraints from responses rather than constraints leading to responses, CRAB aims to reduce noise and ensure precision in training data, often employing Python for verification. 
In contrast, RPO operates by reversing constraints after both constraints and responses have been established, simplifying the process significantly. While CRAB's method can potentially produce constraints of varying quality, the simplicity of RPO's reversal mechanism minimizes the likelihood of errors. 
Furthermore, CRAB is focused on natural language generation (NLG) tasks, requiring large language models (LLMs) to fully understand the constraint system to generate diverse and suitable constraints. RPO differs by targeting discrimination tasks, where the primary effort is in assessing whether constraints are met, and any reversals involve minimal overhead. 
This fundamental difference highlights RPO's efficiency and simplicity in handling constraints compared to CRAB's more complex generation process.





\section{Discussion of Character-driven Data Generation}

\subsection{The relation of Character-driven and real-world scenarios}
We view the relationship between ``character-driven'' and ``real-world scenarios'' in this way: In real-world scenarios, we are essentially engaging in conversations with a variety of characters, such as agents that handle various tasks, role-playing models that take on different roles, or a default LLM (which can also be seen as a character).

Without the introduction of characters, actually, there is also one character introduced, which is a default, generic LLM character.
We consider that if default LLMs are directly applied to a wide range of agent/role-play scenarios, models trained with highly templated constraints may lack generalization capabilities.
Conversely, there are researches (e.g., Persona~\cite{ge2024scaling} and Tulu~\cite{lambert2025tulu3}.) indicate that training LLM on character data can generalize well to general instruction tasks and enable the model to cover a wider range of downstream scenarios.

\subsection{Considerations in data construction}
The introduction of characters is mainly intended to enrich diversity, and characters essentially serve as an inspired corpus~\cite{xu2023wizardlm}. After incorporating characters, the LLM can generate more varied constraints.

Previous methods typically constructed data by concatenating instructions with constraints sampled from a pre-constructed constraint pool. However, the number of constraints in this pool is limited. Besides, considering that these constraints need to pair well with most instructions, they are often highly abstract, atomic, mechanical, and templated, which greatly restricts the expressive range of the constraints.

For instance, the examples from the constraint pool are typically of the following kind, as they can be concatenated to most instruction data without compatibility issues:
\begin{itemize}
    \item Use less than 100 words.
    \item Answer with at least one emoji.
\end{itemize}

However, this method cannot generate richer, more complex, and more realistic constraints like the following, as such constraints must be contextually relevant to specific scenarios and cannot be concatenated to arbitrary instructions:
\begin{itemize}
    \item When a user's symptoms involve a myocardial infarction, the user must be advised to seek help from a legitimate medical institution (medical scenario).
    \item Can only help the user with math problems, cannot answer questions from other subjects, and must politely refuse (educational scenario).
\end{itemize}

After integrating the character, you only need to provide the LLM with background information about the character and a constraints taxonomy, and the LLM can generate various constraints that are compatible with that character. Instead of sampling from a fixed constraint pool, this approach essentially eliminates issues of homogeneity in the generated data since different characters can incentivize LLM to generate various constraints.

It is worth noting that, after introducing characters, we can still generate the atomic constraints mentioned above. Therefore, in our observation, the introduction of character-driven data collection primarily brings benefits.

\section{Discussion about Constraints that Can not be Reversed}
\label{appdix:implicit_constrain}
We observed that at the current stage, the benchmarks for instruction following are primarily explicit and clear constraints for the convenience of evaluation, where each constraint is explicitly extracted.  
However, there indeed exist some constraints that cannot be reversed, as shown in Table~\ref{fig:cannot_be_reversed_example}.

Moreover, regardless of the form in which constraints are presented, for a response and a constraint (which might be directly extracted from the instructions or might need to be rewritten), there must be a clear determination of following/not following. 
Otherwise, the entire evaluation logic of the instruction-following task cannot stand. Considering this, the idea of RPO is makes sense in most scenarios.

Lastly, although RPO cannot handle all possible preferences, it is still capable of addressing a considerable portion and remains valuable. Otherwise, RPO would not be able to achieve a clear advantage over other methods, suggesting that despite the potential issues with ``implicit'' constraints, RPO still has significant advantages over other methods.

\section{More Experimental Detail}

\subsection{Data collection}

\noindent\textbf{Profile data}
We use \textit{GPT-4o-mini} to generate profile generation, detail prompt can be found in Appendix \ref{appdix:profile_prompt}.

\noindent\textbf{SFT data}
The SFT responses generated by \textit{GPT-4o-mini}.
If we cannot successfully obtain a good enough response~(follow rate greater than 0.8) within 5 sample times, 
we perform 5 times refinement on the best response.
For refinement, we provide LLMs with the constraints that need to be followed, the user query, the best response, and the evaluation details about which constraints were successfully followed and which constraints failed to be followed.
If the response is still not good enough after refinement, we end this session.
Through self-play, we collect 2,440 dialogue histories with 11894 turns, each consisting of at most five turns.

\noindent\textbf{DPO/KTO/RPO data}
The responses for DPO/RPO were sampled from the \textit{Llama-3.1 8B Instruct}.
The Instruct model is directly trained on the preference data, without SFT stage.
We have tested the effects of performing alignment after SFT versus direct alignment, and found no significant difference between the two approaches. 
As observed in Table \ref{tab:main_result}, the performance of the SFT model was inferior to that of the Instruct model in some cases. 
We believe this might be due to the fact that the Instruct model has already undergone instruction tuning, especially for latest models (such as \textit{LLaMA-3.1} and \textit{Qwen-2.5}) which have been optimized for instruction-following tasks. Consequently, the benefits of performing SFT are not particularly pronounced.

\noindent\textbf{KTO data}
The response source of KTO is the same as DPO and RPO, just different in using them to construct preference data.
For KTO, we choose to sample the highest-scoring and lowest-scoring responses, using them as positive and negative examples, respectively.
As the label KTO is a binary value~(True/False), to reduce the impact of noise, for the highest-scoring responses, we remove the constraints that it failed to follow in the system prompt. 
Conversely, for the lowest-scoring responses, we remove the constraints that it successfully followed.

\subsection{Hyperparameter experiment of $\gamma$}
Here we provide a detailed analysis of the hyperparameter $\gamma$.
We experiment with six different $\gamma$ values ranging from 0 to 0.2.

\begin{table*}[]
    \centering
    \resizebox{0.8\textwidth}{!}{
    \begin{tabular}{ccccccccc}
        \toprule
    \multirow{2}{*}{\textbf{$\gamma$}} &  \multicolumn{4}{c}{\textbf{Sysbench}}  & \multicolumn{4}{c}{\textbf{Multi-IF}} \\ 
    \cmidrule(lr){2-5} \cmidrule(lr){6-9}
    &  \textbf{CSR} & \textbf{ISR} & \textbf{SSR} & \textbf{Avg.}   & \textbf{Step 1} & \textbf{Step 2} & \textbf{Step 3}  & \textbf{Avg.} \\ 
    \midrule
        \multicolumn{9}{c}{\textbf{Llama-3.1-8B-Instruct RPO}} \\
        \midrule
        0.00 & 81.57 & 69.77 & 45.06 & 65.47 & 76.20 & 68.98 & 59.21 & 68.13 \\
        0.01 & 82.47 & 70.56 & 45.23 & 66.10 & 75.50 & 68.91 & 60.23 & 68.21 \\
        0.05 & 83.10 & 71.27 & 46.99 & 67.12 & 77.50 & 69.47 & 60.57 & 69.18 \\
        0.10 & 81.79 & 69.21 & 45.48 & 65.49 & 76.40 & 69.14 & 61.53 & 69.02 \\
        0.15 & 82.07 & 70.21 & 44.19 & 65.49 & 75.64 & 68.43 & 59.65 & 67.90 \\
        0.20 & 81.76 & 69.68 & 43.74 & 65.06 & 77.18 & 70.07 & 61.62 & 69.62 \\
        \midrule
        \multicolumn{9}{c}{\textbf{Qwen-2.5-7B-Instruct RPO}} \\
        \midrule
        0.00 & 78.01 & 65.58 & 41.10 & 61.56 & 76.81 & 63.99 & 55.57 & 65.46 \\
        0.01 & 79.92 & 66.58 & 41.48 & 62.66 & 77.06 & 65.83 & 56.68 & 66.52 \\
        0.05 & 80.25 & 67.16 & 41.77 & 63.06  & 77.25 & 65.43 & 56.12 & 66.27 \\
        0.10 & 81.82 & 69.00 & 44.40 & 65.07 & 77.34 & 65.31 & 55.95 & 66.20 \\
        0.15 & 81.24 & 68.37 & 43.13 & 64.25 & 77.22 & 64.78 & 55.49 & 65.83 \\
        0.20 & 81.56 & 68.50 & 44.31 & 64.79 & 77.19 & 64.89 & 56.10 & 66.06 \\
        \bottomrule
    \end{tabular}
    }
    \caption{Performance of different $\gamma$.}
\end{table*}

\subsection{Error Analyze}
\label{appdix:error_analyze}
We sampled 500 constraints along with their reversed versions from our training data and manually analyzed their quality.
The annotators are requested to categorize the data into four types. The proportions of the four types are as follows:

\begin{table}[h]
    \centering
    \begin{tabular}{lc}
        \toprule
        \textbf{Type} & \textbf{Percentage} \\
        \midrule
        Perfect reversal & 85.2 \% \\
        Slight error reversal (still acceptable) & 4.0 \% \\
        Bad reversal (unacceptable) & 8.4 \% \\
        The Constrain can not be reversed & 2.4 \% \\
        \bottomrule
    \end{tabular}
    \caption{Percentages of different reversal types.}
\end{table}

Here are some cases of these four types:

\begin{table*}[h]
    \centering
    \small
    \begin{tabular}{p{0.48\linewidth} p{0.48\linewidth}}
        \toprule
        \textbf{Original Constraint} & \textbf{Reversed Constraint (Predicted)} \\
        \midrule
        In any case, the response length must not exceed 200 words. & In any case, the response length must exceed 200 words. \\
        In all responses, a rating prompt must be included at the end. & In all responses, a rating prompt must not be included at the end. \\
        When the user mentions the weather, the response must include a weather-related joke. & When the user mentions the weather, the response must not include a weather-related joke. \\
        \bottomrule
    \end{tabular}
    \caption{Example of perfect reversal.}
\end{table*}

\begin{table*}[h]
    \centering
    \small
    \begin{tabular}{p{0.26\linewidth} p{0.34\linewidth} p{0.34\linewidth}}
        \toprule
        \textbf{Original Constraint} & \textbf{Reversed Constraint (Predicted)} & \textbf{Reversed Constraint (Golden)} \\
        \midrule
        When responding, you must use a professional tone and avoid using slang or overly colloquial expressions. & When responding, you can use a casual tone and are encouraged to use slang or colloquial expressions. & When responding, you can use a non-professional tone and are encouraged to use slang or colloquial expressions. \\
        If the user mentions quantum mechanics, it must be explained using metaphors. & If the user mentions quantum mechanics, it must be explained straightforwardly, without using metaphors. & If the user mentions quantum mechanics, metaphors must not be used in the explanation. \\
        No complete sentences are allowed in any response; response must only consist of phrases or keywords. & Complete sentences must appear in any response. & Any response must contain complete sentences and should not consist solely of phrases or keywords. \\
        \bottomrule
    \end{tabular}
    \caption{Example of slight error reversal (still acceptable).}
\end{table*}

\begin{table*}[h]
    \centering
    \small
    \begin{tabular}{p{0.26\linewidth} p{0.34\linewidth} p{0.34\linewidth}}
        \toprule
        \textbf{Original Constraint} & \textbf{Reversed Constraint (Predicted)} & \textbf{Reversed Constraint (Golden)} \\
        \midrule
        Replies must end with an exclamation mark. & Replies must end with a period. & Replies must not end with an exclamation mark. \\
        When the user mentions 'challenge', they must be encouraged to keep going. & When the user mentions 'challenge', they must be discouraged from continuing. & When the user mentions 'challenge', there is no need to encourage them to keep going. \\
        Regardless of the topic, each reply must end with a question. & Regardless of the topic, each reply must end with a statement. & \\
        \bottomrule
    \end{tabular}
    \caption{Example of bad reversal (unacceptable).}
\end{table*}

\begin{table*}[h]
    \centering
    \small
    \begin{tabular}{p{0.98\linewidth}} 
        \toprule
        \textbf{Original Constraint}   \\
        \midrule 
        The target audience is high school girls.\\
        When the user guesses incorrectly three consecutive times, the answer will then be provided. \\
        Grade the answer based on the following criteria: 1-2 points for deviation; 3-5 points for being average; 6-10 points for accuracy and strong logic. \\
        \bottomrule
    \end{tabular}
    \label{fig:cannot_be_reversed_example}
    \caption{Example of constraints that can not be reversed .}
\end{table*}

\clearpage

\section{Detail of Reversing Constraint}
\label{appdix:reverse_constraint_prompt}

\small
\begin{longtable}{p{16cm}} 
\toprule
You are an expert in language understanding. I will provide you with some constraint preferences, and I need you to transform these constraints into their opposite meanings.\\
\\
For example, you can make changes in two ways. Please choose the more natural approach based on the actual situation, or combine both:\\
\\
Change "must/should/need to xxx" to "cannot/forbidden to xxx", and change "cannot/forbidden to xxx" to "must xxx".\\
Keep words like "must/cannot/should/strictly forbidden" but change the conditions they need to meet. For example, change "must use the same number of exclamation marks" to "must use different numbers of exclamation marks".\\
In other words, given a real response that doesn't meet the original constraint, I want you to reverse the constraint itself so that this response, which doesn't satisfy the original constraint, fully satisfies the new reversed constraint.\\
\\
Please note that your reversed constraints should cover exactly all opposing situations to the original constraint, not just some situations that don't meet the original constraint. Otherwise, a response that doesn't meet the original constraint might still not meet the new reversed constraint. You need to pay special attention to cases involving numbers, as shown in Rule 1 below.\\
\\
Please pay particular attention to these situations:\\
\\
1. For constraints involving numbers, for example, if the original constraint is "at least three profession-related keywords must appear", the opposite should be "at most three times" rather than "only once" or "at most twice", as these don't cover all situations opposite to the original constraint.\\
2. Your reversed constraints should be explicit constraints. For example, if the original constraint is "forbidden to mention Van Gogh", while its opposite is "can mention Van Gogh", this isn't explicit enough. I can't measure whether a response satisfies this reversed constraint because it's ambiguous - it can either be satisfied or not. An explicit reversed constraint should be "must mention Van Gogh".\\
I will give you a list of constraints, and you must return them strictly in the following format, without outputting anything else:\\
\\
\texttt{[}\\
\hspace{1em}\{'original\_cons': original constraint 1, 'reverse\_cons': reversed constraint 1\},\\
\hspace{1em}\{'original\_cons': original constraint 2, 'reverse\_cons': reversed constraint 2\},\\
...\\
\texttt{]}\\\\
\\
Here is the list of constraints you need to reverse: \\

        \bottomrule 
    \caption{Prompt of reverse constraint.}
    \label{tab:reverse_constraint_prompt}

\end{longtable}

\begin{table*}[h]
    \centering
    \resizebox{0.99\textwidth}{!}{
    \begin{tabular}{l}
        \toprule
        \textbf{Original constraint:} In any case, the response length must not exceed 200 words. \\
        \textbf{Reversed constraint(Predicted):} In any case, the response length must not be less than 200 words. \\
        \midrule
        \textbf{Original constraint:} In any feedback request, provide at least three reasons for the star rating. \\
        \textbf{Reversed constraint(Predicted):} In any feedback request, provide at most three reasons for the star rating. \\
        \midrule
        \textbf{Original constraint:} In any response, if 'apology' or similar terms are detected, delete that part of the content directly. \\
        \textbf{Reversed constraint(Predicted):} In any response, if 'apology' or similar terms are detected, that part of the content must be retained. \\
        \midrule
        \textbf{Original constraint:} In all responses, a rating prompt must be included at the end. \\
        \textbf{Reversed constraint(Predicted):} In all responses, a rating prompt must not be included at the end. \\
        \midrule
        \textbf{Original constraint:} Before formally answering, clarify the topic so that users understand the rating criteria. \\
        \textbf{Reversed constraint(Predicted):} Before formally answering, the topic must be vague so that users do not understand the rating criteria. \\
        \midrule
        \textbf{Original constraint:} When the user input is empty, automatically repeat the role's prompt information. \\
        \textbf{Reversed constraint(Predicted):} When the user input is empty, repeating the role's prompt information is absolutely not allowed. \\
        \bottomrule
    \end{tabular}
    }
    \caption{Reverse constraint examples.}
    \label{tab:reverse_constraint_example}
\end{table*}

\begin{table*}[h]
    \centering
    \resizebox{0.99\textwidth}{!}{
    \begin{tabular}{l}
        \toprule
        \textbf{Original constraint:} Replies must end with an exclamation mark. \\
        \textbf{Reversed constraint(Predicted):} Replies must end with a period. \\
        \textbf{Reversed constraint(Golden):} Replies must not end with an exclamation mark. \\
        \midrule
        \textbf{Original constraint:} When discussing product features, each reply must include at least one metaphor. \\
        \textbf{Reversed constraint(Predicted):} When discussing product features, each reply must include at least one specific description. \\
        \textbf{Reversed constraint(Golden):} When discussing product features, each reply must not include metaphors. \\
        \midrule
        \textbf{Original constraint:} Regardless of the topic, each reply must end with a question. \\
        \textbf{Reversed constraint(Predicted):} Regardless of the topic, each reply must end with a statement. \\
        \textbf{Reversed constraint(Golden):} Regardless of the topic, each reply must not end with a question. \\
        \midrule
        \textbf{Original constraint:} Use at least one metaphor in reply. \\
        \textbf{Reversed constraint(Predicted):} Use a maximum of one metaphor in reply. \\
        \textbf{Reversed constraint(Golden):} Do not use metaphors in reply. \\
        \midrule
        \textbf{Original constraint:} When dealing with calculations, results should be represented in Chinese uppercase numerals. \\
        \textbf{Reversed constraint(Predicted):} When dealing with calculations, results must be represented in Arabic numerals. \\
        \textbf{Reversed constraint(Golden):} When dealing with calculations, results should be represented in a form other than Chinese uppercase numerals. \\
        \midrule
        \textbf{Original constraint:} If the user mentions their failed attempts, encouragement must be given. \\
        \textbf{Reversed constraint(Predicted):} If the user mentions their failed attempts, criticism must be given. \\
        \textbf{Reversed constraint(Golden):} If the user mentions their failed attempts, do not give encouragement. \\
        \midrule
        \textbf{Original constraint:} When providing regular expressions, emojis must be added at the end. \\
        \textbf{Reversed constraint(Predicted):} When providing regular expressions, any emojis must be removed at the end. \\
        \textbf{Reversed constraint(Golden):} When providing regular expressions, adding emojis at the end is prohibited. \\
        \bottomrule
    \end{tabular}
    }
    \caption{Bad cases of reverse constraint.}
    \label{tab: bad_case}
\end{table*}

\clearpage

\section{Prompt of constraint Ontology}
\label{appdix:constraint_ontology_prompt}

\small
\begin{longtable}{p{16cm}}
\toprule
1. Content:\\
\hspace{2em} - `What content should be included in responses`: For example, keywords, emojis, punctuation, numbers, content from input, quotes from poems, idioms, jokes, examples, etc.\\
\hspace{2em} - When certain content is mentioned, what must or must not be mentioned.\\
\hspace{2em} - `Number of times specific content should be included`: For example, at least three times.\\
\hspace{2em}   - `Content that must not be mentioned in responses`: For example, prohibiting certain keywords, prohibiting the use of emojis, commas, Arabic numerals, prohibiting quoting poems, idioms, prohibiting discussion on certain topics.\\
   \hspace{2em}- `Starting content` (used in a minority of cases, less than 15\%): For example, starting with "Hello," a catchphrase, a specific line of poetry, self-introduction.\\
   \hspace{2em}- `Ending content` (used in a minority of cases, less than 15\%): For example, ending with "Yo," "~," "Good luck," emojis, "Bingo!", a quote from someone, a specific sentence, etc.\\
   \hspace{2em}- `Very special preferences` (extremely rare, less than 5\%, please combine with actual scenarios): Including but not limited to using spaces between every character, repeating the user's input first, using uppercase Chinese numerals for numbers, etc.\\
   \hspace{2em}- ...\\
\\
2. Format:\\
   \hspace{2em}- `Overall format`: For example, returning in JSON format, markdown format, expressing in paragraphs, returning according to a given template.\\
   \hspace{2em}- `Partial format`: For example, if multiple steps are involved, they should be answered point by point; titles should be highlighted using \#\#; enclosed in angle brackets <>; certain parts should be bolded using \*\*bold\*\*, or italicized using \*italic\*.\\
   \hspace{2em}- `Length`: For example, limiting response length to exceed/not exceed a certain number of characters.\\
   \hspace{2em}- `Numerical limits`: Limiting the number of sentences, sections, key points mentioned, suggestions given, recommended movies, etc.\\
  \hspace{2em} - ...\\
\\
3. Style:\\
   \hspace{2em}- `Tone`: For example, responses must be professional/informal/angry/sarcastic/impatient/vintage style/cute/rugged/gentlemanly... (\*\*The constraints added here must not conflict with any possible character setting in the given character reference information, so please add such constraints cautiously\*\*).\\
   \hspace{2em}- `Audience`: For example, explaining as if speaking to a three-year-old child.\\
   \hspace{2em}- `Writing style`: For example, poetic style, novel style, sonnet style, classical Chinese style, Wikipedia entry style, Xiaohongshu style, flowery language style, etc.\\
   \hspace{2em}- `Figures of speech`: For example, using metaphors, rhetorical questions, etc.\\
   \hspace{2em}- `Language`: For example, responses must be in Chinese/English, which parts in Chinese, which parts in English, which parts should be in French, limitations on uppercase and lowercase letters in English (initials), some parts returned in phonetic symbols or pinyin, etc.\\
   \hspace{2em}- ...\\
\\
4. Role (used in very few cases, less than 5\%, please combine with actual scenarios, use cautiously):\\
   \hspace{2em}- `Identity`: For example, each response must first declare who they are, etc.\\
   \hspace{2em}- `Self-reference`: For example, must refer to oneself as "we," or "this system," etc.\\
   \hspace{2em}- `Narration`: For example, responses must use parentheses () to indicate body movements, etc.\\
   \hspace{2em}- ...\\
\\
5. Logic (in some cases, please combine with actual scenarios, use cautiously):\\
   \hspace{2em}- For example, for problems involving calculations, the answer must be increased by one, etc.\\
   - ...\\
\\
6. Behavior/Actions:\\
   \hspace{2em}- `Refusal`: For example, when the user mentions certain topics such as politics, topics unrelated to skills, robot privacy, offensive questions, etc., refuse to answer or provide suggestions, or say "I don't know," or steer the conversation back, etc.\\
   \hspace{2em}- `Invitation`: For example, when the user shows interest in certain things/mentions certain things, invite them to further discuss the topic.\\
   \hspace{2em}- `Clarification`: For example, for unclear instructions from the user, the system may proactively request clarification.\\
   \hspace{2em}- ...\\
        \bottomrule  
    \caption{Constrain ontology, used in system prompt generation.}
    \label{tab:constrain_ontology}
\end{longtable}
\clearpage

\section{Detail of Profile Generation}
\label{appdix:profile_prompt}

\small
\begin{longtable}{p{16cm}}
\toprule
You are a master of character modeling and constraint generation.\\
\\
Here is the basic information about the character (a chatbot) you need to model:\\
<Start of Character Reference Information>\\
\{system\_inspired\_corpus\}\\
<End of Character Reference Information>\\
\\
Please refer to the information above and write a complete Chinese profile for the character.\\
\\
The profile information includes:\\
<Start of Profile Content>\\
1. The name of the character.\\
2. Character description and basic information. Generate a vivid and three-dimensional description of the character based on the given reference information, with no less than 100 characters.\\
3. Skill points, used to provide the user with references on which issues can be discussed with this chatbot.\\
4. Constraints related to the character's description and skills, serving as preferences for the chatbot's responses to user questions. Generate 6 to 12 non-repetitive, non-conflicting, and distinctive constraints as needed.\\
\\
Constraints refer to the preferences a chatbot must meet when answering human questions. Below is an ontology system of sample constraints. The content of the constraints can include, but is not limited to, the following situations. This is not a complete system of constraints:\\
<Start of Constraint System>\\
\{CONSTRAIN\_ONTOLOGY\}\\
\\
In addition to the content of the constraints, these constraints must also have specific triggering scenarios. For example, when introducing Beijing cuisine, Peking duck must be mentioned; when the user expresses offense, a joke needs to be told, etc. The trigger can also be applicable in any situation and not tied to any specific topic or scenario, such as in casual conversation scenarios, for example, all responses must be within 100 characters.\\
<End of Constraint System>\\
\\
<End of Profile Content>\\
\\
Here are some conditions that must be met by the constraints:\\
<Start of Constraint Rules>\\
1. The content of the constraints and their trigger conditions must be \*\*clear\*\* and \*\*objectively verifiable\*\*, not vague or ambiguous. Generate more constraints that can be objectively verified, even constraints that can be directly checked using Python code. Avoid generating constraints that are too subjective, as this increases the difficulty and randomness of evaluation.\\
2. Constraints need not be limited to the given constraint system. The provided system is just a reference and not complete. Use your imagination fully and brainstorm. Encourage thinking outside the given ontology's preference limitations and generate constraints with \*\*high creativity, personality, richness, diversity, and unexpected quirks\*\*.\\
3. Constraints should fit the usage scenario of this chatbot. It is not necessary to generate constraints one by one according to the ontology. Generate constraints suitable for these scenarios based on the specific situation of the character.
4. There must be absolutely no conflict between the generated constraints.
5. Constraints do not necessarily need to have a positive impact on the quality of responses. Do not assume that this must be a helpful, concise, friendly, emotionless, and personality-free robot, as this will limit the proposal of interesting and diverse constraints.\\
6. Constraints do not necessarily need to be simple atomic constraints and can be complex composite constraints.\\
7. The constraints you generate will be used only for \*\*immediate, single-round, pure text dialogue\*\* scenarios. \*\*Absolutely avoid\*\* generating triggers like "if the user is silent for more than 30 seconds," which are \*\*unrelated to pure text dialogue\*\*; \*\*absolutely avoid\*\* generating constraints like "when the user expresses interest in xx for three consecutive rounds," "during the first round of conversation with the user," which require \*\*considering multi-round dialogue context\*\*; \*\*absolutely avoid\*\* generating constraints like "send greetings to the user once a week," which are for \*\*non-immediate dialogue\*\* scenarios.
8. Trigger conditions should preferably not be simply "when the user mentions xx topic." Creative trigger conditions are encouraged.\\
<End of Constraint Rules>\\
\\
You must strictly follow the format below to respond, and do not output any other content:\\
\\ 
\{\\
\hspace{2em}"Name": "The name of this chatbot",\\
\hspace{2em}"Description": "The description of this chatbot",\\
\hspace{2em}"Skill": ["Content of Skill 1", "Content of Skill 2", ...],\\
\hspace{2em}"Constraints": ["Content of Constraint 1", "Content of Constraint 2", "Content of Constraint 3", ...]\\
\}\\
\\
        \bottomrule  
    \caption{Prompt of system profile generation.}
    \label{tab:profile_prompt}
\end{longtable}

\small
\begin{longtable}{p{16cm}}
\toprule
You are an expert in character modeling. I will provide you with basic information about a character. \\
Please generate a vivid and three-dimensional profile for this character, which can cover basic information, personality traits, the way they communicate/ask questions, tone of voice, background story, inner beliefs, etc. \\
Please use your imagination to create a character image that is unique and lively, and avoid creating generic characters. \\
You don't have to make all aspects of the character positive; if necessary, you can include some neutral or slightly negative traits to increase diversity.\\
You only need to return the profile you generate and do not output any other content.\\
The basic information of the character is as follows:\\
\{BASIC\_INFO\}\\
\bottomrule
    \caption{Prompt of user profile generation.}
    \label{tab:user_profile_prompt}
\end{longtable}

 

\small
\begin{longtable}{p{16cm}}
\toprule
\textbf{Original user profile from Persona}\\
A middle-aged professional looking to reconnect with their Italian heritage through learning traditional recipes from the grandmother \\
\midrule
\textbf{Our extended version}\\
\*\*Character Profile: Marco Rossi\*\*\\
\\
\*\*Basic Information:\*\*\\
- Name: Marco Rossi\\
- Age: 45\\
- Occupation: Marketing Manager\\
- Residence: San Francisco, California, USA\\
- Nationality: Italian-American\\
\\
\*\*Personality Traits:\*\*\\
Marco presents himself as a mature and composed individual, often dressed in smart suits that give him a professional appearance. However, deep down, he harbors a zest for life. He has a strong yearning for family traditions, particularly Italian culinary culture. There is a contradiction in his personality: he is a decisive decision-maker at work, yet appears somewhat hesitant and sensitive in familial and emotional matters. He cares deeply for his friends and family and often feels anxious about their happiness.\\
\\
\*\*Communication Style:\*\*\\
Marco communicates with a gentle and engaging tone. He enjoys using vivid metaphors to express his views and often quotes Italian proverbs. His voice is deep and magnetic, with a hint of an Italian accent, especially when he talks about food, where his tone naturally rises, revealing his passion. His catchphrase is: "Life is like a bowl of pasta; the choice of seasoning determines its flavor."\\
\\
\*\*Background Story:\*\*\\
Marco was born into an Italian immigrant family, with his grandmother being the closest person to him during his childhood. Every weekend, he would visit her, listening to stories about Italy and learning to make traditional Italian dishes. However, as time went by, Marco became engulfed in his busy work schedule and modern life, and his connection with his grandmother dwindled. After her passing, he felt a deep sense of loss, realizing that his connection to Italian culture was gradually fading.\\
\\
In an effort to reconnect with Italy, Marco decided to learn his grandmother's traditional recipes. He began attending cooking classes, trying to recreate the delicious meals he had enjoyed as a child. His kitchen is filled with the aroma of spices, and friends often gather at his home to share in his culinary discoveries. He hopes to preserve his grandmother's memory through food and to introduce more people to the charm of Italian culture.\\
\\
\*\*Inner Beliefs:\*\*\\
Marco believes that food is not just a necessity for survival but also a vessel for culture and emotion. He firmly believes that family and tradition are the most important aspects of life. He often reminds himself: "No matter how busy life gets, remembering your roots is the key to finding your true self." Through his efforts, he hopes to pass on Italian traditional cuisine and culture to the next generation, allowing them to feel the warmth and love from their ancestors.\\
\bottomrule
    \caption{Example of original user profile from Persona~\cite{ge2024scaling} and the final user profile after we expand it.}
    \label{tab:user_profile_case}
\end{longtable}

\small
\begin{longtable}{p{16cm}}
\toprule
\textbf{Original GPTs from GPT-store}\\
Blame Shifter Master is a chatbot specialized in handling workplace blame-shifting issues. Whether a colleague or a boss tries to shift responsibility onto you, Blame Shifter Master offers effective strategies and techniques to help you gracefully navigate the situation while maintaining professionalism. It is your best ally in workplace struggles.\\
\midrule
\textbf{Our revised version}\\ 

\# Profile:\\
\*\*Name\*\*: Blame Shifter Master\\
\*\*Description\*\*: Blame Shifter Master is a chatbot specialized in handling workplace blame-shifting issues. Whether a colleague or a boss tries to shift responsibility onto you, Blame Shifter Master offers effective strategies and techniques to help you gracefully navigate the situation while maintaining professionalism. It is your best ally in workplace struggles.\\
\\
\# Skills:\\
Provides practical strategies to avoid being blamed in the workplace
Teaches how to elegantly counteract unfair treatment at work\\
Shares tips on building a strong professional image to reduce the likelihood of being targeted\\
Analyzes specific cases and offers personalized responses\\
\\
\# Constraints:\\
1. Whenever the user mentions 'boss' or 'supervisor', the response must use 'wise decision-maker' at least once to maintain a positive and respectful tone.\\
2. In any response that mentions a solution, it must begin with 'First, stay calm; second, ...' to emphasize the importance of composure.
3. Negative words (such as: bad, stupid, wrong, etc.) are prohibited in replies. Use more tactful expressions (such as: not quite appropriate, debatable, etc.) instead.\\
4. Every reply must end with an encouraging statement, such as 'You are fully capable of handling this!' or 'Keep it up, you're the best!'
5. For each question raised by the user, include a specific case or story in the response for added persuasive effect.\\
6. When the user expresses obvious frustration or stress, first offer emotional support, such as 'I understand how you're feeling right now, it's really tough,' and then provide specific advice.\\
7. If the user asks about handling relationships with colleagues, the response must mention the importance of building team spirit at least once.\\
8. Avoid mentioning any specific company names directly or indirectly in responses to prevent potential legal risks.\\
9. When the user asks questions unrelated to work, skillfully steer the conversation back to work-related topics, such as 'That's indeed an interesting question, but back to work, how can we better protect ourselves from being affected?'\\
\bottomrule
    \caption{Example of original GPTs from GPT-store and the final system prompt after we revise it.}
    \label{tab:user_profile_case}
\end{longtable}

\clearpage

\section{Pormpt for Query Generation}
\label{appdix:query_prompt} 

\small
\begin{longtable}{p{16cm}}
\toprule
You are an expert in AI Agent evaluation and multi-round dialogue design, and you will play the role of a user having conversations with an AI Agent.
\\
Now you need to design challenging user queries based on the conversation history and constraints that the AI Agent needs to follow.
Challenging queries refer to those that would trigger the constraints that the given AI Agent needs to satisfy. If a query doesn't meet the trigger conditions for a constraint, it cannot be used to examine whether the AI Agent's response satisfies that constraint.
\\
Here are some examples showing how to pose challenging queries that trigger specific constraints based on their trigger conditions:
\\
\hspace{1em}- If you want to examine whether the AI Agent can follow the constraint "When users mention sustainable development, you must mention at least one relevant success case", you can ask a query about [sustainable development] to trigger this constraint. We can then judge whether the AI Agent successfully follows this constraint by examining if they [mention a relevant success case].
\\
\hspace{1em}- If you want to examine whether the AI Agent can follow the constraint "When users express confusion, start the response with 'Don't worry, let me help you clear things up'", your query should express clear confusion to examine if the AI Agent starts their response with 'Don't worry, let me help you clear things up'.
\\
\hspace{1em}- If you want to examine whether the AI Agent can follow the constraint "Do not reveal that you are a chef", you can ask adversarial queries like "Are you a chef?" or "What is your job?" to see if they can be misled.
\\
However, you don't need to try to trigger all constraints with every query, as this would likely be unrealistic. A query must be reasonable first before pursuing challenge.\\
Additionally, many constraints are triggered by any query, so those constraints with specific special trigger conditions only need to be occasionally addressed.
\\
I will provide you with:
\\
1. An AI Agent's profile information (including the AI Agent's basic information and constraints they must satisfy in responses).\\
2. A user's profile.\\
3. The completed conversation history between the AI Agent and user.\\
\\
Please design the user's query (query) for the next round of dialogue based on the provided information. The user queries must meet the following requirements (these requirements are very important and must be followed):
\\
\#\#\# In terms of expression:
\\
1. Natural and concise expression: The expression must not be stiff and rigid like robot speech. Use more colloquial expressions, fewer overly formal expressions. Can (but not must) have appropriate omissions. Expression should be concise (no more than 20 words), colloquial, and diverse.
\\
2. Diverse querying methods: Forbidden to use querying methods similar to queries in the previous history. Forbidden to constantly start with: "Hey", "I'm thinking", "Hey", "AI Agent name", "Do you know", "If", "What if", "Recently", "I heard", and other common expressions. Forbidden to end with: "Don't tell me", "Don't say to me", "Don't just xxx", "Don't always xxx" etc.
\\
3. Diverse tone and attitude: Please use your imagination to create queries with as diverse tones as possible. You can choose any tone to generate queries, including but not limited to: {characteristic}, etc., or combinations of these tones (but tones cannot conflict with each other).
\\
4. Consistent persona: Expression should be consistent with the user modeling in the user profile I provided, conforming to the user's gender, age, background, occupation, tone, personality traits, communication style, and other settings in the user profile.
\\
5. Do not use AI Agent's full name: If the AI Agent's name is too formal or too long, such as: Rail Fence Cipher Master, Mathematical Superhero (Second Generation). You cannot directly copy their name exactly to address them, as that's too mechanical and unlike normal human expression. You can appropriately abbreviate the address, use suitable nicknames, or directly use "you". I don't want every query to start with the AI Agent's name.\\
6. Do not use emojis.\\
\\
\#\#\# In terms of query content:
1. Diverse query content: Please fully utilize your imagination and creativity. You can ask very nonsensical queries, and have very diverse, imaginative, unexpected queries and ideas.
2. Chat topics: User queries don't necessarily have to relate to the domains and skills mentioned in the AI Agent's profile. There can be a 10\% chance of generating casual chat queries.\\
\\
\#\#\# In terms of key examination content:
\\
1. Generated queries should be challenging enough, focusing on generating queries that will trigger the constraints this AI Agent needs to satisfy or queries that have continuity related to previous conversation history.
\\
2. Continuity doesn't mean rigidly following up on elements in the AI Agent's response, but reasonably continuing the dialogue following previous history. Don't use rigid ways like "Since you just mentioned xx is interesting" to establish connections with historical dialogue.
\\
3. Generated queries can relate to content from any previous round of dialogue, not necessarily just the last query.
\\
4. Continuity is not mandatory and must be based on reasonable, natural premises. If it's difficult to produce naturally continuous queries, you can generate queries without strong connections to previous dialogue history.\\
\\
Your output should only be the user query itself, without any other content.
\\
Here is the information provided to you:
\\
<AI\_Agent\_profile\_start>\\
THE\_SYSTEM\_PROFILE\\
<AI\_Agent\_profile\_end>\\

<user\_profile\_start>\\
THE\_USER\_PROFILE\\
<user\_profile\_end>\\

<conversation\_history\_start>\\
THE\_CONVERSATION\_HISTORY\\
<conversation\_history\_end>\\

Next round's user query:\\
        \bottomrule
    \caption{Prompt for query generation.}
    \label{tab:query_prompt}
\end{longtable}

\clearpage

\section{Prompt for Fine-grained Evaluation}
\label{appdix:evaluation_prompt}

\small
\begin{longtable}{p{15cm}}
        \toprule
        You are now an expert in evaluating the results of large models. Below, you will face an evaluation task regarding a large model's ability to adhere to a system prompt.
        \\
        I will provide you with the current round's query, the current round's response, and the evaluation criteria. You need to accurately assess and inform the adherence status of each constraint in the evaluation criteria.
        Current Round's Query:
        \\
        <user\_query\_start>\\
        THE\_USER\_QUERY\\
        <user\_query\_end>\\
        \\
        Current Round's Response:\\
        <response\_start>\\
        THE\_USER\_RESPONSE\\
        <response\_end>\\
        \\
        Evaluation Criteria Details\\
        <constraint\_list\_start>\\
        THE\_constraint\_LIST\\
        <constraint\_list\_end>\\
        \\
        Please carefully read all the constraints in the above evaluation criteria and strictly use them as the standard for judgment. For each requirement in the evaluation criteria, please assess whether the current round's response adheres to them one by one, in order.
        \\
        In very rare cases, a particular evaluation criterion may not apply to the current response. For example, if the evaluation criterion is "When describing AI application scenarios, at least one metaphor must be used," but the response completely lacks any mention of AI application scenarios, then it cannot be determined whether this response meets this evaluation criterion. In such cases, the evaluation result for this criterion will be True.
        \\
        Please use this rule with extreme caution; such situations are very rare, and the evaluation criteria are carefully written. Only use it when you are 100\% sure.
        \\
        You must strictly return in the following JSON format, containing three fields: constraint, reasoning for the judgment, and the evaluation result (which can only be True or False).
        \\
        The output format is as follows:\\
        \texttt{[}\\
            \hspace{1em}\{\\
                \hspace{2em}"constraint": "This is the constraint being judged, it must be copied exactly", \\
                \hspace{2em}"reasoning": "Why you think this response meets/does not meet the first constraint", \\
                \hspace{2em}"evaluation\_result": True or False\\
            \hspace{1em}\},\\
            \hspace{1em}\{\\
                \hspace{2em}"constraint": "...", \\
                \hspace{2em}"reasoning": "...",\\
                \hspace{2em}"evaluation\_result": ...\\
            \hspace{2em}\},\\
            ...\\
        \texttt{]}\\
        
        \bottomrule 
    \caption{Prompt of fine-grained evaluation.}
    \label{tab:evaluation_prompt}
\end{longtable}

%% file: acl_latex.bbl
\begin{thebibliography}{34}
\providecommand{\natexlab}[1]{#1}

\bibitem[{Bradley and Terry(1952)}]{Bradley1952RankAO}
Ralph~Allan Bradley and Milton~E. Terry. 1952.
\newblock Rank analysis of incomplete block designs: I. the method of paired comparisons.
\newblock \emph{Biometrika}, 39:324.

\bibitem[{Brown et~al.(2020)Brown, Mann, Ryder, Subbiah, Kaplan, Dhariwal, Neelakantan, Shyam, Sastry, Askell, Agarwal, Herbert-Voss, Krueger, Henighan, Child, Ramesh, Ziegler, Wu, Winter, Hesse, Chen, Sigler, Litwin, Gray, Chess, Clark, Berner, McCandlish, Radford, Sutskever, and Amodei}]{brown2023language}
Tom Brown, Benjamin Mann, Nick Ryder, Melanie Subbiah, Jared~D Kaplan, Prafulla Dhariwal, Arvind Neelakantan, Pranav Shyam, Girish Sastry, Amanda Askell, Sandhini Agarwal, Ariel Herbert-Voss, Gretchen Krueger, Tom Henighan, Rewon Child, Aditya Ramesh, Daniel Ziegler, Jeffrey Wu, Clemens Winter, Chris Hesse, Mark Chen, Eric Sigler, Mateusz Litwin, Scott Gray, Benjamin Chess, Jack Clark, Christopher Berner, Sam McCandlish, Alec Radford, Ilya Sutskever, and Dario Amodei. 2020.
\newblock \href {https://proceedings.neurips.cc/paper_files/paper/2020/file/1457c0d6bfcb4967418bfb8ac142f64a-Paper.pdf} {Language models are few-shot learners}.
\newblock In \emph{Advances in Neural Information Processing Systems}, volume~33, pages 1877--1901. Curran Associates, Inc.

\bibitem[{Chen et~al.(2021)Chen, Tworek, Jun, Yuan, de~Oliveira~Pinto, Kaplan, Edwards, Burda, Joseph, Brockman, Ray, Puri, Krueger, Petrov, Khlaaf, Sastry, Mishkin, Chan, Gray, Ryder, Pavlov, Power, Kaiser, Bavarian, Winter, Tillet, Such, Cummings, Plappert, Chantzis, Barnes, Herbert-Voss, Guss, Nichol, Paino, Tezak, Tang, Babuschkin, Balaji, Jain, Saunders, Hesse, Carr, Leike, Achiam, Misra, Morikawa, Radford, Knight, Brundage, Murati, Mayer, Welinder, McGrew, Amodei, McCandlish, Sutskever, and Zaremba}]{chen2021codex}
Mark Chen, Jerry Tworek, Heewoo Jun, Qiming Yuan, Henrique~Ponde de~Oliveira~Pinto, Jared Kaplan, Harri Edwards, Yuri Burda, Nicholas Joseph, Greg Brockman, Alex Ray, Raul Puri, Gretchen Krueger, Michael Petrov, Heidy Khlaaf, Girish Sastry, Pamela Mishkin, Brooke Chan, Scott Gray, Nick Ryder, Mikhail Pavlov, Alethea Power, Lukasz Kaiser, Mohammad Bavarian, Clemens Winter, Philippe Tillet, Felipe~Petroski Such, Dave Cummings, Matthias Plappert, Fotios Chantzis, Elizabeth Barnes, Ariel Herbert-Voss, William~Hebgen Guss, Alex Nichol, Alex Paino, Nikolas Tezak, Jie Tang, Igor Babuschkin, Suchir Balaji, Shantanu Jain, William Saunders, Christopher Hesse, Andrew~N. Carr, Jan Leike, Josh Achiam, Vedant Misra, Evan Morikawa, Alec Radford, Matthew Knight, Miles Brundage, Mira Murati, Katie Mayer, Peter Welinder, Bob McGrew, Dario Amodei, Sam McCandlish, Ilya Sutskever, and Wojciech Zaremba. 2021.
\newblock \href {https://arxiv.org/abs/2107.03374} {Evaluating large language models trained on code}.

\bibitem[{Cheng et~al.(2024{\natexlab{a}})Cheng, Liu, Wang, Gu, Lu, Zhang, Dong, Tang, Wang, and Huang}]{cheng2024spar}
Jiale Cheng, Xiao Liu, Cunxiang Wang, Xiaotao Gu, Yida Lu, Dan Zhang, Yuxiao Dong, Jie Tang, Hongning Wang, and Minlie Huang. 2024{\natexlab{a}}.
\newblock \href {https://arxiv.org/abs/2412.11605} {Spar: Self-play with tree-search refinement to improve instruction-following in large language models}.
\newblock \emph{Preprint}, arXiv:2412.11605.

\bibitem[{Cheng et~al.(2024{\natexlab{b}})Cheng, Lu, Gu, Ke, Liu, Dong, Wang, Tang, and Huang}]{cheng2024autodetect}
Jiale Cheng, Yida Lu, Xiaotao Gu, Pei Ke, Xiao Liu, Yuxiao Dong, Hongning Wang, Jie Tang, and Minlie Huang. 2024{\natexlab{b}}.
\newblock Autodetect: Towards a unified framework for automated weakness detection in large language models.
\newblock \emph{arXiv preprint arXiv:2406.16714}.

\bibitem[{Cobbe et~al.(2021)Cobbe, Kosaraju, Bavarian, Chen, Jun, Kaiser, Plappert, Tworek, Hilton, Nakano, Hesse, and Schulman}]{cobbe2021gsm8k}
Karl Cobbe, Vineet Kosaraju, Mohammad Bavarian, Mark Chen, Heewoo Jun, Lukasz Kaiser, Matthias Plappert, Jerry Tworek, Jacob Hilton, Reiichiro Nakano, Christopher Hesse, and John Schulman. 2021.
\newblock Training verifiers to solve math word problems.
\newblock \emph{arXiv preprint arXiv:2110.14168}.

\bibitem[{Dong et~al.(2024)Dong, Lu, Li, Xia, Yu, Zhou, and Zhou}]{dong2024self}
Guanting Dong, Keming Lu, Chengpeng Li, Tingyu Xia, Bowen Yu, Chang Zhou, and Jingren Zhou. 2024.
\newblock Self-play with execution feedback: Improving instruction-following capabilities of large language models.
\newblock \emph{arXiv preprint arXiv:2406.13542}.

\bibitem[{Du et~al.(2022)Du, Qian, Liu, Ding, Qiu, Yang, and Tang}]{du2022glmgenerallanguagemodel}
Zhengxiao Du, Yujie Qian, Xiao Liu, Ming Ding, Jiezhong Qiu, Zhilin Yang, and Jie Tang. 2022.
\newblock \href {https://arxiv.org/abs/2103.10360} {Glm: General language model pretraining with autoregressive blank infilling}.
\newblock \emph{Preprint}, arXiv:2103.10360.

\bibitem[{Ethayarajh et~al.(2024)Ethayarajh, Xu, Muennighoff, Jurafsky, and Kiela}]{ethayarajh2024kto}
Kawin Ethayarajh, Winnie Xu, Niklas Muennighoff, Dan Jurafsky, and Douwe Kiela. 2024.
\newblock \href {https://arxiv.org/abs/2402.01306} {Kto: Model alignment as prospect theoretic optimization}.
\newblock \emph{Preprint}, arXiv:2402.01306.

\bibitem[{Ge et~al.(2024)Ge, Chan, Wang, Yu, Mi, and Yu}]{ge2024scaling}
Tao Ge, Xin Chan, Xiaoyang Wang, Dian Yu, Haitao Mi, and Dong Yu. 2024.
\newblock \href {https://arxiv.org/abs/2406.20094} {Scaling synthetic data creation with 1,000,000,000 personas}.
\newblock \emph{Preprint}, arXiv:2406.20094.

\bibitem[{Guo et~al.(2024)Guo, Zhang, Liu, Liu, Khalman, Llinares, Rame, Mesnard, Zhao, Piot, Ferret, and Blondel}]{guo2024direct}
Shangmin Guo, Biao Zhang, Tianlin Liu, Tianqi Liu, Misha Khalman, Felipe Llinares, Alexandre Rame, Thomas Mesnard, Yao Zhao, Bilal Piot, Johan Ferret, and Mathieu Blondel. 2024.
\newblock \href {https://arxiv.org/abs/2402.04792} {Direct language model alignment from online ai feedback}.
\newblock \emph{Preprint}, arXiv:2402.04792.

\bibitem[{Hacohen and Weinshall(2019)}]{hacohen2019power}
Guy Hacohen and Daphna Weinshall. 2019.
\newblock \href {https://arxiv.org/abs/1904.03626} {On the power of curriculum learning in training deep networks}.
\newblock \emph{Preprint}, arXiv:1904.03626.

\bibitem[{He et~al.(2024{\natexlab{a}})He, Zeng, He, Liang, and Xiao}]{he2024complex}
Qianyu He, Jie Zeng, Qianxi He, Jiaqing Liang, and Yanghua Xiao. 2024{\natexlab{a}}.
\newblock \href {https://arxiv.org/abs/2404.15846} {From complex to simple: Enhancing multi-constraint complex instruction following ability of large language models}.
\newblock \emph{Preprint}, arXiv:2404.15846.

\bibitem[{He et~al.(2024{\natexlab{b}})He, Jin, Wang, Bi, Mandyam, Zhang, Zhu, Li, Xu, Lv et~al.}]{he2024multi}
Yun He, Di~Jin, Chaoqi Wang, Chloe Bi, Karishma Mandyam, Hejia Zhang, Chen Zhu, Ning Li, Tengyu Xu, Hongjiang Lv, et~al. 2024{\natexlab{b}}.
\newblock Multi-if: Benchmarking llms on multi-turn and multilingual instructions following.
\newblock \emph{arXiv preprint arXiv:2410.15553}.

\bibitem[{Hu et~al.(2022)Hu, Shen, Wallis, Allen-Zhu, Li, Wang, Wang, and Chen}]{hu2022lora}
Edward~J Hu, Yelong Shen, Phillip Wallis, Zeyuan Allen-Zhu, Yuanzhi Li, Shean Wang, Lu~Wang, and Weizhu Chen. 2022.
\newblock \href {https://openreview.net/forum?id=nZeVKeeFYf9} {Lo{RA}: Low-rank adaptation of large language models}.
\newblock In \emph{International Conference on Learning Representations}.

\bibitem[{Jiang et~al.(2024{\natexlab{a}})Jiang, Sablayrolles, Roux, Mensch, Savary, Bamford, Chaplot, de~las Casas, Hanna, Bressand, Lengyel, Bour, Lample, Lavaud, Saulnier, Lachaux, Stock, Subramanian, Yang, Antoniak, Scao, Gervet, Lavril, Wang, Lacroix, and Sayed}]{jiang2024mixtralexperts}
Albert~Q. Jiang, Alexandre Sablayrolles, Antoine Roux, Arthur Mensch, Blanche Savary, Chris Bamford, Devendra~Singh Chaplot, Diego de~las Casas, Emma~Bou Hanna, Florian Bressand, Gianna Lengyel, Guillaume Bour, Guillaume Lample, Lélio~Renard Lavaud, Lucile Saulnier, Marie-Anne Lachaux, Pierre Stock, Sandeep Subramanian, Sophia Yang, Szymon Antoniak, Teven~Le Scao, Théophile Gervet, Thibaut Lavril, Thomas Wang, Timothée Lacroix, and William~El Sayed. 2024{\natexlab{a}}.
\newblock \href {https://arxiv.org/abs/2401.04088} {Mixtral of experts}.
\newblock \emph{Preprint}, arXiv:2401.04088.

\bibitem[{Jiang et~al.(2024{\natexlab{b}})Jiang, Wang, Zeng, Zhong, Li, Mi, Shang, Jiang, Liu, and Wang}]{jiang2024followbench}
Yuxin Jiang, Yufei Wang, Xingshan Zeng, Wanjun Zhong, Liangyou Li, Fei Mi, Lifeng Shang, Xin Jiang, Qun Liu, and Wei Wang. 2024{\natexlab{b}}.
\newblock \href {https://aclanthology.org/2024.acl-long.257} {{F}ollow{B}ench: A multi-level fine-grained constraints following benchmark for large language models}.
\newblock In \emph{Proceedings of the 62nd Annual Meeting of the Association for Computational Linguistics (Volume 1: Long Papers)}, pages 4667--4688, Bangkok, Thailand. Association for Computational Linguistics.

\bibitem[{Lambert et~al.(2025)Lambert, Morrison, Pyatkin, Huang, Ivison, Brahman, Miranda, Liu, Dziri, Lyu, Gu, Malik, Graf, Hwang, Yang, Bras, Tafjord, Wilhelm, Soldaini, Smith, Wang, Dasigi, and Hajishirzi}]{lambert2025tulu3}
Nathan Lambert, Jacob Morrison, Valentina Pyatkin, Shengyi Huang, Hamish Ivison, Faeze Brahman, Lester James~V. Miranda, Alisa Liu, Nouha Dziri, Shane Lyu, Yuling Gu, Saumya Malik, Victoria Graf, Jena~D. Hwang, Jiangjiang Yang, Ronan~Le Bras, Oyvind Tafjord, Chris Wilhelm, Luca Soldaini, Noah~A. Smith, Yizhong Wang, Pradeep Dasigi, and Hannaneh Hajishirzi. 2025.
\newblock \href {https://arxiv.org/abs/2411.15124} {Tulu 3: Pushing frontiers in open language model post-training}.
\newblock \emph{Preprint}, arXiv:2411.15124.

\bibitem[{Liu et~al.(2025)Liu, Fang, Hu, Zhang, Zhou, Zhang, Tu, Lin, Huang, Song et~al.}]{liu2025survey}
Shunyu Liu, Wenkai Fang, Zetian Hu, Junjie Zhang, Yang Zhou, Kongcheng Zhang, Rongcheng Tu, Ting-En Lin, Fei Huang, Mingli Song, et~al. 2025.
\newblock A survey of direct preference optimization.
\newblock \emph{arXiv preprint arXiv:2503.11701}.

\bibitem[{Liu et~al.(2024)Liu, Lei, Wang, Huang, Feng, Wen, Cheng, Ke, Xu, Tam, Zhang, Sun, Gu, Wang, Zhang, Huang, Dong, and Tang}]{liu2024alignbench}
Xiao Liu, Xuanyu Lei, Shengyuan Wang, Yue Huang, Zhuoer Feng, Bosi Wen, Jiale Cheng, Pei Ke, Yifan Xu, Weng~Lam Tam, Xiaohan Zhang, Lichao Sun, Xiaotao Gu, Hongning Wang, Jing Zhang, Minlie Huang, Yuxiao Dong, and Jie Tang. 2024.
\newblock \href {https://arxiv.org/abs/2311.18743} {Alignbench: Benchmarking chinese alignment of large language models}.
\newblock \emph{Preprint}, arXiv:2311.18743.

\bibitem[{OpenAI(2023)}]{openai2023gpt4}
OpenAI. 2023.
\newblock \href {https://arxiv.org/abs/2303.08774} {{GPT-4 Technical Report}}.
\newblock \emph{Preprint}, arXiv:2303.08774.

\bibitem[{Ouyang et~al.(2022)Ouyang, Wu, Jiang, Almeida, Wainwright, Mishkin, Zhang, Agarwal, Slama, Ray, Schulman, Hilton, Kelton, Miller, Simens, Askell, Welinder, Christiano, Leike, and Lowe}]{ouyang2022rlhf}
Long Ouyang, Jeff Wu, Xu~Jiang, Diogo Almeida, Carroll~L. Wainwright, Pamela Mishkin, Chong Zhang, Sandhini Agarwal, Katarina Slama, Alex Ray, John Schulman, Jacob Hilton, Fraser Kelton, Luke Miller, Maddie Simens, Amanda Askell, Peter Welinder, Paul Christiano, Jan Leike, and Ryan Lowe. 2022.
\newblock \href {https://arxiv.org/abs/2203.02155} {Training language models to follow instructions with human feedback}.
\newblock \emph{Preprint}, arXiv:2203.02155.

\bibitem[{Qi et~al.(2024)Qi, Peng, Wang, Xu, Hou, and Li}]{qi2024constraint}
Yunjia Qi, Hao Peng, Xiaozhi Wang, Bin Xu, Lei Hou, and Juanzi Li. 2024.
\newblock \href {https://arxiv.org/abs/2410.24175} {Constraint back-translation improves complex instruction following of large language models}.
\newblock \emph{Preprint}, arXiv:2410.24175.

\bibitem[{Qin et~al.(2024{\natexlab{a}})Qin, Zhang, Shen, Luo, Sun, Zhang, Qiao, Chen, Zhou, Zhang et~al.}]{qin2024sysbench}
Yanzhao Qin, Tao Zhang, Yanjun Shen, Wenjing Luo, Haoze Sun, Yan Zhang, Yujing Qiao, Weipeng Chen, Zenan Zhou, Wentao Zhang, et~al. 2024{\natexlab{a}}.
\newblock Sysbench: Can large language models follow system messages?
\newblock \emph{arXiv preprint arXiv:2408.10943}.

\bibitem[{Qin et~al.(2024{\natexlab{b}})Qin, Song, Hu, Yao, Cho, Wang, Wu, Liu, Liu, and Yu}]{qin2024infobench}
Yiwei Qin, Kaiqiang Song, Yebowen Hu, Wenlin Yao, Sangwoo Cho, Xiaoyang Wang, Xuansheng Wu, Fei Liu, Pengfei Liu, and Dong Yu. 2024{\natexlab{b}}.
\newblock Infobench: Evaluating instruction following ability in large language models.
\newblock \emph{arxiv preprint arXiv:2401.03601}.

\bibitem[{Rafailov et~al.(2024)Rafailov, Sharma, Mitchell, Ermon, Manning, and Finn}]{rafailov2024direct}
Rafael Rafailov, Archit Sharma, Eric Mitchell, Stefano Ermon, Christopher~D. Manning, and Chelsea Finn. 2024.
\newblock \href {https://arxiv.org/abs/2305.18290} {Direct preference optimization: Your language model is secretly a reward model}.
\newblock \emph{Preprint}, arXiv:2305.18290.

\bibitem[{Sun et~al.(2024)Sun, Liu, Li, Wang, Dong, Lin, and Huang}]{sun2024conifer}
Haoran Sun, Lixin Liu, Junjie Li, Fengyu Wang, Baohua Dong, Ran Lin, and Ruohui Huang. 2024.
\newblock \href {https://arxiv.org/abs/2404.02823} {Conifer: Improving complex constrained instruction-following ability of large language models}.
\newblock \emph{arxiv preprint arXiv:2404.02823}.

\bibitem[{Tao et~al.(2024)Tao, Lin, Chen, Li, Wu, Li, Jin, Huang, Tao, and Zhou}]{tao2024survey}
Zhengwei Tao, Ting-En Lin, Xiancai Chen, Hangyu Li, Yuchuan Wu, Yongbin Li, Zhi Jin, Fei Huang, Dacheng Tao, and Jingren Zhou. 2024.
\newblock A survey on self-evolution of large language models.
\newblock \emph{arXiv preprint arXiv:2404.14387}.

\bibitem[{Wen et~al.(2024)Wen, Ke, Gu, Wu, Huang, Zhou, Li, Hu, Gao, Xu et~al.}]{wen2024benchmarking}
Bosi Wen, Pei Ke, Xiaotao Gu, Lindong Wu, Hao Huang, Jinfeng Zhou, Wenchuang Li, Binxin Hu, Wendy Gao, Jiaxin Xu, et~al. 2024.
\newblock Benchmarking complex instruction-following with multiple constraints composition.
\newblock \emph{arXiv preprint arXiv:2407.03978}.

\bibitem[{Xu et~al.(2023)Xu, Sun, Zheng, Geng, Zhao, Feng, Tao, and Jiang}]{xu2023wizardlm}
Can Xu, Qingfeng Sun, Kai Zheng, Xiubo Geng, Pu~Zhao, Jiazhan Feng, Chongyang Tao, and Daxin Jiang. 2023.
\newblock \href {https://arxiv.org/abs/2304.12244} {Wizardlm: Empowering large language models to follow complex instructions}.
\newblock \emph{Preprint}, arXiv:2304.12244.

\bibitem[{Zhang et~al.(2024{\natexlab{a}})Zhang, Shen, Luo, Zhang, Liang, Yang, Lin, Qiao, Chen, Cui et~al.}]{zhang2024cfbench}
Tao Zhang, Yanjun Shen, Wenjing Luo, Yan Zhang, Hao Liang, Fan Yang, Mingan Lin, Yujing Qiao, Weipeng Chen, Bin Cui, et~al. 2024{\natexlab{a}}.
\newblock Cfbench: A comprehensive constraints-following benchmark for llms.
\newblock \emph{arXiv preprint arXiv:2408.01122}.

\bibitem[{Zhang et~al.(2024{\natexlab{b}})Zhang, Yu, Fu, Huang, and Li}]{zhang2024iopo}
Xinghua Zhang, Haiyang Yu, Cheng Fu, Fei Huang, and Yongbin Li. 2024{\natexlab{b}}.
\newblock Iopo: Empowering llms with complex instruction following via input-output preference optimization.
\newblock \emph{arXiv preprint arXiv:2411.06208}.

\bibitem[{Zheng et~al.(2024)Zheng, Zhang, Zhang, Ye, Luo, Feng, and Ma}]{zheng2024llamafactory}
Yaowei Zheng, Richong Zhang, Junhao Zhang, Yanhan Ye, Zheyan Luo, Zhangchi Feng, and Yongqiang Ma. 2024.
\newblock \href {http://arxiv.org/abs/2403.13372} {Llamafactory: Unified efficient fine-tuning of 100+ language models}.
\newblock In \emph{Proceedings of the 62nd Annual Meeting of the Association for Computational Linguistics (Volume 3: System Demonstrations)}, Bangkok, Thailand. Association for Computational Linguistics.

\bibitem[{Zhou et~al.(2023)Zhou, Lu, Mishra, Brahma, Basu, Luan, Zhou, and Hou}]{zhou2023instruction}
Jeffrey Zhou, Tianjian Lu, Swaroop Mishra, Siddhartha Brahma, Sujoy Basu, Yi~Luan, Denny Zhou, and Le~Hou. 2023.
\newblock Instruction-following evaluation for large language models.
\newblock \emph{arXiv preprint arXiv:2311.07911}.

\end{thebibliography}
